%% file: main.tex
\documentclass{article}

\usepackage[final]{style}

\usepackage[utf8]{inputenc} 
\usepackage[T1]{fontenc}    
\usepackage{hyperref}       
\usepackage{url}            
\usepackage{booktabs}       
\usepackage{amsfonts}       
\usepackage{nicefrac}       
\usepackage{microtype}      
\usepackage{wrapfig}
\usepackage{xcolor}

\usepackage{graphicx}
\usepackage{caption}
\usepackage{subcaption}
\usepackage{tablefootnote}
\usepackage{footnote}
\usepackage[inline]{enumitem}
\usepackage{xurl}
\usepackage{tcolorbox}

\renewcommand{\paragraph}[1]{\vspace{-5pt}\smallskip\par\textbf{#1.}}

\title{The Benchmark Lottery}

\author{%
 Mostafa Dehghani\thanks{Equal contribution, $^\dagger$equal advising.} \\ 
 Google Brain \\ 
  \texttt{dehghani@google.com} \\
  \And
  Yi Tay$^*$ \\ 
  Google Research \\ 
  \texttt{yitay@google.com} \\
  \And 
  Alexey A. Gritsenko$^*$ \\ 
  Google Brain\\ 
  \texttt{agritsenko@google.com} \\ 
  \And 
  Zhe Zhao\\ 
  Google Brain\\ 
  \texttt{zhezhao@google.com} \\ 
  \And 
  Neil Houlsby\\ 
  Google Brain\\ 
  \texttt{neilhoulsby@google.com} \\ 
  \And 
  Fernando Diaz\\ 
  Google Brain \\ 
  \small{\texttt{diazfernando@google.com}} \\ 
  \And 
  Donald Metzler$^\dagger$\\ 
  Google Research\\ 
  \texttt{metzler@google.com} \\ 
  \And 
  Oriol Vinyals$^\dagger$\\ 
  DeepMind \\ 
  \texttt{vinyals@google.com} \\ %
}

\begin{document}
\maketitle

\begin{abstract}
The world of empirical machine learning (ML) strongly relies on benchmarks in order to determine the relative effectiveness of different algorithms and methods. This paper proposes the notion of \emph{a benchmark lottery} that describes the overall fragility of the ML benchmarking process. The benchmark lottery postulates that many factors, other than fundamental algorithmic superiority, may lead to a method being perceived as superior. On multiple benchmark setups that are prevalent in the ML community, we show that the relative performance of algorithms may be altered significantly simply by choosing different benchmark tasks, highlighting the fragility of the current paradigms and potential fallacious interpretation derived from benchmarking ML methods. Given that every benchmark makes a statement about what it perceives to be important, we argue that this might lead to biased progress in the community. We discuss the implications of the observed phenomena and provide recommendations on mitigating them using multiple machine learning domains and communities as use cases, including natural language processing, computer vision, information retrieval, recommender systems, and reinforcement learning.
\end{abstract}
\tableofcontents

\section{Introduction}
Quantitative evaluation is a cornerstone of machine learning research.  As a result, benchmarks, including those based on data sets and simulations, have become fundamental to tracking the progress of machine learning research. 

Benchmarks have a long history in artificial intelligence research generally. There have been several attempts at designing milestones to capture progress toward artificial intelligence (e.g., human level game performance, the  Turing test~\citep{Turing1950}).  Specific system properties are measured through specialized benchmarks (e.g. for vision, natural language processing, robotics).  All of these benchmarks, by design, encode values about what is salient and important, both across domains (e.g. natural language processing benchmarks versus robotics benchmarks) and within them (e.g. which languages are considered in an NLP benchmark, which environments are considered in a robotics benchmark).

As benchmarks become widely accepted, researchers adopt them, often without questioning their assumptions, and algorithmic development becomes slowly tied to these success metrics. Indeed, over time, the research community makes collective decisions about what shared tasks--and values--are important (through peer review norms and resource investment) and which are not.

Because of this, it is important for the research community to understand the individual, community, social, and political pressures that influence why some benchmarks become canonical and others do not.
This paper shares some opinions on this topic along with case studies calling for discussion and reconsiderations on several issues with benchmarking in machine learning and argues that a meta-level understanding of benchmarks is a prerequisite for understanding how the progress in machine learning is made.  

This paper presents analyses on how benchmarks may affect the direction and pace of progress in machine learning and puts forward the notion of a benchmark lottery.  We argue that many factors other than the algorithmic superiority of a method may influence the emergence of algorithms that are perceived as better. Moreover, we claim that for a method to emerge successful, it has to first win the \emph{benchmark lottery}. Out of the many potential trials in this lottery, a method has to be first well-aligned with the suite of benchmarks that the community has accepted as canonical. We refer to the alignment between the tasks brought forth by the community and successful algorithms as the \emph{task selection} bias. We empirically show that the task selection process has a great influence over the relative performance of different methods. Moreover, we argue that benchmarks are \emph{stateful}, meaning that the method has to also participate in the lottery at the right moment, and to align well with existing techniques, tricks, and state-of-the-art. Related to this, we also briefly discuss how benchmark reuse may affect the statistical validity of the results of new methods. 

As a whole, as we researchers continue to participate in the benchmark lottery, there are long-term implications, which we believe are important to be explicitly aware of. As such, the main goals of this paper are to
\begin{enumerate*}[label=(\roman*)]
  \item raise awareness of these phenomena and potential issues they create; and to, 
  \item provide some recommendations for mitigating these issues. 
\end{enumerate*}
We argue that community forces and task selection biases, if left unchecked, may lead to unwarranted overemphasis of certain types of models and to unfairly hinder the growth of other classes of models - which may be important for making fast and reliable progress in machine learning. 


Additionally, we discuss the trend of consolidating several benchmarks into a meta-benchmark (or benchmark suite), e.g. GLUE~\citep{wang-etal-2018-glue}, XTREME~\citep{hu2020xtreme}, VTAB~\citep{zhai2019large}, RL Unplugged~\citep{gulcehre2020rl} to name a few. In this context, using the most common score aggregation technique, i.e., averaging, a model that performs poorly on \emph{any} of the bundled tasks may be set up for failure. 
We argue that the \emph{and} operator is an inductive bias that favors well-rounded models as opposed to models that may do really well on one task, making unfair assumptions about what practitioners might really need or want. Finally, in alignment with the task selection biases, we show that the selection of a subset of the tasks used in the benchmark suite plays a substantial role in determining the winning algorithms, providing further evidence that researchers are participating in the benchmark lottery. 

The notion of what makes a benchmark canonical, in the sense that is widely accepted by the community, is also diverse depending on the field of study. On one hand, fields like natural language processing (NLP) or computer vision (CV) have well-established benchmarks for certain problems. On the other hand, fields such as recommender systems or reinforcement learning tend to allow researchers more freedom in choosing their own tasks and evaluation criteria for comparing methods. We show how this may act as \emph{rigging the lottery}, where researchers can ``make their own luck'' by fitting benchmarks and experimental setups to models instead.

Overall, this paper explores these aspects of model evaluation in machine learning research. We frame this from a new perspective of the \emph{benchmark lottery}. While there has been recent work that peers deeply into the benchmark tasks themselves~\citep{bowman2021will}, this work takes meta- and macro-perspectives to encompass factors that go beyond designing reliable standalone tasks.

The remainder of the paper is organized as follow:
\begin{itemize}
    \item Section~\ref{section:background} discusses how benchmarks can influence long-term research directions in a given (sub-)field, and describes the life cycle of a benchmark. 
    \item Section~\ref{section:task-selection-bias} introduces the \emph{task selection bias} and using established benchmarks as examples shows how relative performance of algorithms is affected by the task selection process. 
    \item Section~\ref{section:community_bias} takes another view of the task selection bias and proposes \emph{community bias} as a higher-level process that influences task selection. We show that forces from the broader research community directly impact the task selection process and as a result, play a substantial role in creating the lottery.
    \item Section~\ref{section:reusing-benchmark} posits that benchmarks are stateful entities and that participation in a benchmark differs vastly depending upon its state. We also argue continual re-use of the same benchmark may be problematic.
    \item Section~\ref{section:rigging-the-lottery} discusses \emph{rigging the lottery}, the issue that some communities (e.g. recommender systems and reinforcement learning) face, where the lack of well-established community-driven sets of benchmarks or clear guidelines may inadvertently enable researchers to fit benchmarks to model. We highlight the potential drawbacks of such an approach. 
    \item Finally, in Section~\ref{section:what_can_we_do} we provide recommendations for finding a way out of the lottery by building better benchmarks and rendering more accurate judgments when comparing models.
\end{itemize}

Overall, unified benchmarks have led to incredible progress and breakthroughs in machine learning and artificial intelligence research~\citep{kingma2013auto,mikolov2013distributed,sutskever2014sequence,bahdanau2014neural,goodfellow2014generative,hinton2015distilling,silver2016mastering,he2016deep,vaswani2017attention,devlin2018bert,brown2020language,dosovitskiy2020image}. There is certainly a lot of benefits of having the community come together to solve shared tasks and benchmarks. Given that the role of benchmarks is indispensable and highly important for measuring progress, this work seeks to examine, introspect and find ways to improve. 

\section{Background}
\label{section:background}
Measuring progress is one of the most difficult aspects of empirical computer science and machine learning. Such questions as 
``What are the best setup and task to use for evaluation?''~\citep{ponce2006dataset, machado2018revisiting, lin2019neural, bowman2021will, recht2019imagenet, lin2021significant, gulcehre2020rl, perazzi2016benchmark, vania2020comparing, musgrave2020metric},
``Which data or benchmark are most applicable?''~\citep{metzler2012experimental, beyer2020we, northcutt2021pervasive, gulcehre2020rl, dacrema2019we}, 
``Which metrics are suitable?'' ~\citep{machado2018revisiting, Bouthillier2021, balduzzi2018re, bouthillier2019unreproducible, musgrave2020metric}, 
or ``What are the best practices for fair benchmarking?''~\citep{torralba2011unbiased, armstrong2009improvements, machado2018revisiting, sculley2018winner, lin2019neural, bowman2021will, Bouthillier2021, recht2019imagenet, lin2021significant, balduzzi2018re, lipton2018troubling, bouthillier2019unreproducible, vania2020comparing, mishra2021robust, benjaminccientific, dodge2019show}  
are of utmost importance to correct empirical evaluation of new ideas and algorithms, and have been extensively studied. Nevertheless, the jury is still out on most of these questions.

Owing to this complexity at no point in time can the claim of superiority of a certain algorithm or model on a specific task be made with full confidence. However, providing any answer to the question of ``Which algorithm is currently the best?'' can play a major role in determining the direction of research and the progress in the field in the long run.

We argue that some models and algorithms are not inherently superior to their alternatives, but are instead perceived as such by the research community due to various factors that we discuss in this paper. One of these factors is the software and hardware support for an idea, as captured in the concept of hardware lottery by \citet{hooker2020hardware}. Here however we focus mainly on \emph{benchmarking}-related factors, and discuss the role they play in the selection of a model as ``fashionable'' in the research world, and how this is often conflated with the model being better. 
When a class of models or algorithms gets recognition in the community, there will be more follow up research, adaption to more setups, more tuning and discovery of better configurations, which lead to better results. This is a valid way of propelling the field further. However, a question that we should also ask is how much progress could have been made by investing the same amount of time, effort, computational resources and talent in a different class of models.
In other words, assuming model development as a complex high-dimensional optimization process, in which researchers are exploring a fitness surface, the initial point, as well as the fitness function, are the key factors for ending up with better optima, and both these factors are highly affected by the benchmarks used for evaluation.

Before diving into various aspects of the benchmark lottery, we begin by describing the life cycle of a benchmark and set the stage by describing several widely adopted benchmarks/tasks and their characteristics.

\subsection{Life of a benchmark}

\label{section:life-of-a-benchmark}
A benchmark is often proposed as a means to evaluate relative model quality. However, it can sometimes also be employed as a diagnostic tool or suite that allows practitioners to peer into the behavior of their models. This section examines the life cycle of a benchmark.
\paragraph{Inception}  A benchmark may be first proposed as an example task in its debut paper (implicit), or be directly framed as a shared task (explicit) where the main contribution may be the dataset itself. A recent trend is also for a leaderboard or competition website to be set up to allow submissions from the community. An example of the former is the Stanford Sentiment Treebank (SST) dataset, which was proposed as part of a methods paper~\citep{socher-etal-2013-recursive}. Conversely, examples of an explicit shared task are benchmarks like SQuAD~\citep{rajpurkar2016squad}, GLUE~\citep{wang-etal-2018-glue,wang2019superglue} or XTREME~\citep{hu2020xtreme}.
\paragraph{Gaining traction \emph{or} fizzling out} certain benchmarks may gain traction at an early stage, encouraging a large number of submissions or publications that evaluate their ideas against them. Benchmarks that are gamified with leaderboards~\citep{rajpurkar2016squad, bowman2015large,hu2020xtreme, wang2019superglue,wang-etal-2018-glue} tend to be of this nature, encouraging dataset creators to always setup these competition scenarios. Similar to how posts become viral on social media, early tractions and citations on benchmark are likely to influence future adoption. That said, benchmarks may also fizzle out due to one or many issues. Oversaturation on a certain domain is of the leading factors, along with usage difficulties (such as having evaluation metrics or setups that may be inconvenient or tedious, etc). On the flip side, benchmarks can suddenly gain traction if a highly influential method paper uses them in its experiments. As an example, the subject-verb agreement task~\citep{linzen2016assessing} became more popular, when it started being used for showcasing the ability of recurrent neural networks over Transformers on modeling hierarchical structure~\citep{tran-etal-2018-importance, universaltrans, abnar2020transferring}.
\paragraph{Active and in-progress} This stage may take a long time depending on how much progress is being made on the proposed task. An active task will typically have a large number of submissions and publications attempting to make improvements on the task. Progress may be optimistic and the bar for improvement is usually higher at this stage. A Stage~3 benchmark may even be considered \emph{``hot''} where any progress or state-of-the-art result is poised to generate a lot of interest and publicity. An example of such a benchmark is the Long Range Arena benchmark~\citep{tay2020long}, whose usage continues to increase leading to the emergence of new ideas for improving the efficiency of Transformer models on long sequences. 

\paragraph{Maturity, degeneration or deprecation} A benchmark may reach its maturity if the community considers the task to be solved, or if meaningful progress on the task can no longer be made (degeneration). This can happen once human parity is achieved (e.g. SQuAD). Eventually, the community largely stops paying attention to the task, and what is left is researchers optimizing for the last $0.01\%$ percentage improvement. An example of continued hill-climbing on a mature benchmark can be found at the SQuAD leaderboard\footnote{https://rajpurkar.github.io/SQuAD-explorer/}.

\paragraph{Renewal} Once a benchmark degenerates or reaches full maturity, it is not uncommon for the original authors (or an independent set of authors) to propose new variations or extensions of the benchmark. This might be to resolve issues with the existing version, or simply to improve the challenge, difficulty, or other aspects of the benchmark. We saw this evolution when SQuAD benchmark graduated to V2~\citep{rajpurkar2018know}, or when the SNLI~\citep{bowman2015large} benchmark slowly evolved into MultiNLI~\citep{williams2017broad}. Another example is the of ImageNet-A and O~\citep{hendrycks2021nae}, ImageNet-C~\citep{hendrycks2019robustness}, and ImageNet-R~\citep{hendrycks2020many} for evaluating robustness of algorithms.
As a more known example, QMNIST~\citep{yadav2019cold} is proposed as a larger and more accurate dataset as a replacement for the MNIST dataset.

\section{Task selection bias}
\label{section:task-selection-bias}
As we show in this section, relative model performance is highly sensitive to the choice of tasks and datasets it is measured on. As a result, the selection of well-established benchmarks plays a more important role than is perhaps acknowledged, and constitutes a form of partiality and bias - the \emph{task selection bias}. 

\subsection{Case Studies}
In this section, we study four different popular benchmarks or natural language processing, computer vision, efficient sequence modeling, and offline reinforcement learning and use the data from the leaderboards of these benchmarks and run analyses to highlight the effect of task selection bias. 


\subsubsection{SuperGLUE}
\begin{figure}[t]
    \centering
    \includegraphics[width=0.8\textwidth]{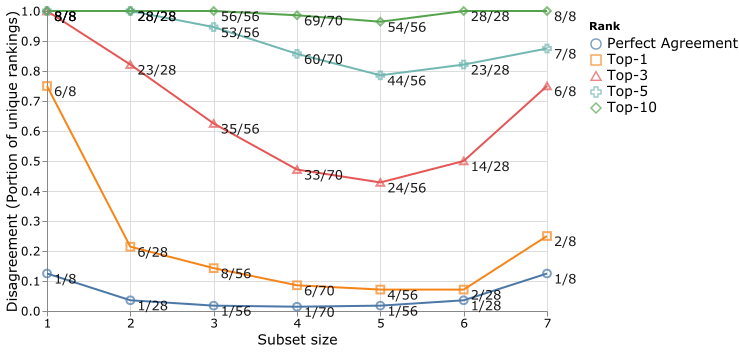}
    \caption{Disagreement of model rankings on the SuperGLUE benchmark as a function of the number of selected benchmark tasks. The $x$-axis represents the number of tasks in each sub-selection of tasks and each line corresponds to a different value of $k$ for the Top-$k$ in the rankings. Points are labels as $A / B$, where $A$ is the number of unique model rankings and $B$ is the total number of possible task combinations for this subset size. If $A=1$, then all rankings are equivalent and consistent across all task selections; higher values of $A$ correspond to higher degrees of disagreement between models rankings. 
    }
    \label{fig:superglue_unique_rankings}
\end{figure}

In order to study the effect of aggregated scores and how findings change by emphasizing and de-emphasizing certain tasks, we explore the SuperGLUE dataset~\citep{wang2019superglue}. To demonstrate the task selection bias on this benchmark, we re-compute the aggregated scores using different combinations of eight SuperGLUE tasks. 

\paragraph{Model Runs} We consider over $55$ different top performing models that are studied in~\citep{narang2021transformer}, including transformer-based models with various activation functions, normalization and parameter initialization schemes, and also architectural extensions (e.g., Evolved Transformers~\citep{so2019evolved}, Synthesizers~\citep{tay2020synthesizer}, Universal Transformer~\citep{universaltrans}, and Switch Transformers~\citep{fedus2021switch}) as well as convolution-based models (e.g. lightweight and dynamic convolutions). We consider the fine-grained scores of these models on the $8$ individual tasks of SuperGLUE and their different combinations. For each combination of tasks, we take a mean-aggregate model performance for all models on the selected tasks and produce a ranking of all $55$ models. To make this ranking more meaningful, we only consider its Top-$k$ entries, where $k \in \{1,3,5,10\}$. 

\paragraph{Ranking inconsistency}
Figure~\ref{fig:superglue_unique_rankings} gives a concise overview of the number of unique Top-$k$ rankings produced obtained from fixed-size subsets of tasks. For example among the $70$ different possibilities of selecting $4$ out of $8$ tasks, there are $6$ distinct model ranking orders produced for Top-$1$ (i.e. there are $6$ different possible top models). Moreover, when considering Top-$3$ or even Top-$5$, almost $60$ out of $70$ rankings do not agree with each other. Overall, the rankings become highly diverse as the subset of tasks selected from the benchmark is varied. This forms the core of the empirical evidence of the task selection bias. 
More analyses on ranking of models on all possible combinations of tasks, rank correlation between SuperGLUE score and individual tasks, effect of relative raking of models in Appendix~\ref{app:superglue_comb},\ref{app:superglue_rank_corr}, and  \ref{app:superglue_rel_rank}.

\subsubsection{Visual Task Adaptation Benchmark (VTAB)}
A similar situation can be observed for the Visual Task Adaptation Benchmark (VTAB; \citep{zhai2019large})  benchmark. VTAB is used for evaluating the quality of representations learned by different models in terms of their ability to adapt to diverse, unseen tasks with few examples. In addition to standard natural image tasks, like classification on ImageNet or CIFAR datasets,  VTAB includes tasks that are related to sensorimotor control, medical imaging, and scene understanding. The benchmark defines the score of an algorithm as its expected performance over a known distribution of tasks that includes those that a human can solve, from visual input alone. 

VTAB defines a total of 19 tasks, grouped into three categories: (\emph{i}) \emph{Natural}, which contains natural images captured using standard cameras that represent generic, fine-grained, or abstract objects [\emph{Caltech101}\citep{fei2006one}, \emph{CIFAR100}~\citep{krizhevsky2009learning}, \emph{DTD}~\citep{cimpoi2014describing}, \emph{Flowers102}~\citep{nilsback2008automated}, \emph{Pets}~\citep{parkhi2012cats}, \emph{Sun397}~\citep{xiao2010sun}, and \emph{SVHN}~\citep{netzer2011reading}.]; (\emph{ii}) \emph{Specialized}, which contains images of the world that captured through specialist equipment [Remote sensing: \emph{Resisc45}~\citep{cheng2017remote} and \emph{EuroSAT}~\citep{helber2019eurosat}: aerial images of the Earth captured using satellites or aerial photography; Medical: \emph{Patch Camelyon}~\citep{veeling2018rotation}, metastases detection from microscopy images, and \emph{Diabetic Retinopathy}~\citep{kaggle_retinopathy}, retinopathy classification from fundus images.]; and finally (\emph{iii}) \emph{Structured}, which contains tasks that designed to assess comprehension of the structure of a scene, mostly generated syntactically using simulated environments [\emph{CLEVR}~\citep{johnson2017clevr}: Simple shapes rendered in a 3D scene, with two tasks: counting and depth prediction, \emph{dSprites}~\citep{higgins2016beta}: Simple black-and-white shapes rendered in 2D, with two tasks: location and orientation prediction, \emph{SmallNORB}~\citep{lecun2004learning}: Artificial objects viewed under varying conditions, with two tasks: object azimuth and camera-elevation prediction, \emph{DMLab}~\citep{beattie2016deepmind}: Frames from a rendered 3D maze. The task involves predicting the time for a pre-trained RL agent to navigate to an object, \emph{KITTI}~\citep{geiger2013vision}: frames captured from a car driver's perspective and the task is to predict the depth of the nearest vehicle.]. We have evaluated $32$ different models against all the $19$ VTAB tasks. The difference between models is  on their architectures (e.g. WAE-GAN~\citep{tolstikhin2017wasserstein} vs. VIVI\citep{tschannen2020self}), their sizes (e.g. ResNet-50 vs. ResNet-101~\citep{kolesnikov2019big}), or the dataset they were pre-trained on (e.g. ResNet-50 pretrained on ImageNet-21k vs. ResNet-50 pretrained on JFT~\citep{kolesnikov2019big}). Models we considered in our study are those that are introduced as ``representation learning algorithms'' in \citep{zhai2019large}.

\begin{figure}[h]
     \centering
     \begin{subfigure}[b]{0.25\textwidth}
         \centering
         \includegraphics[width=\textwidth]{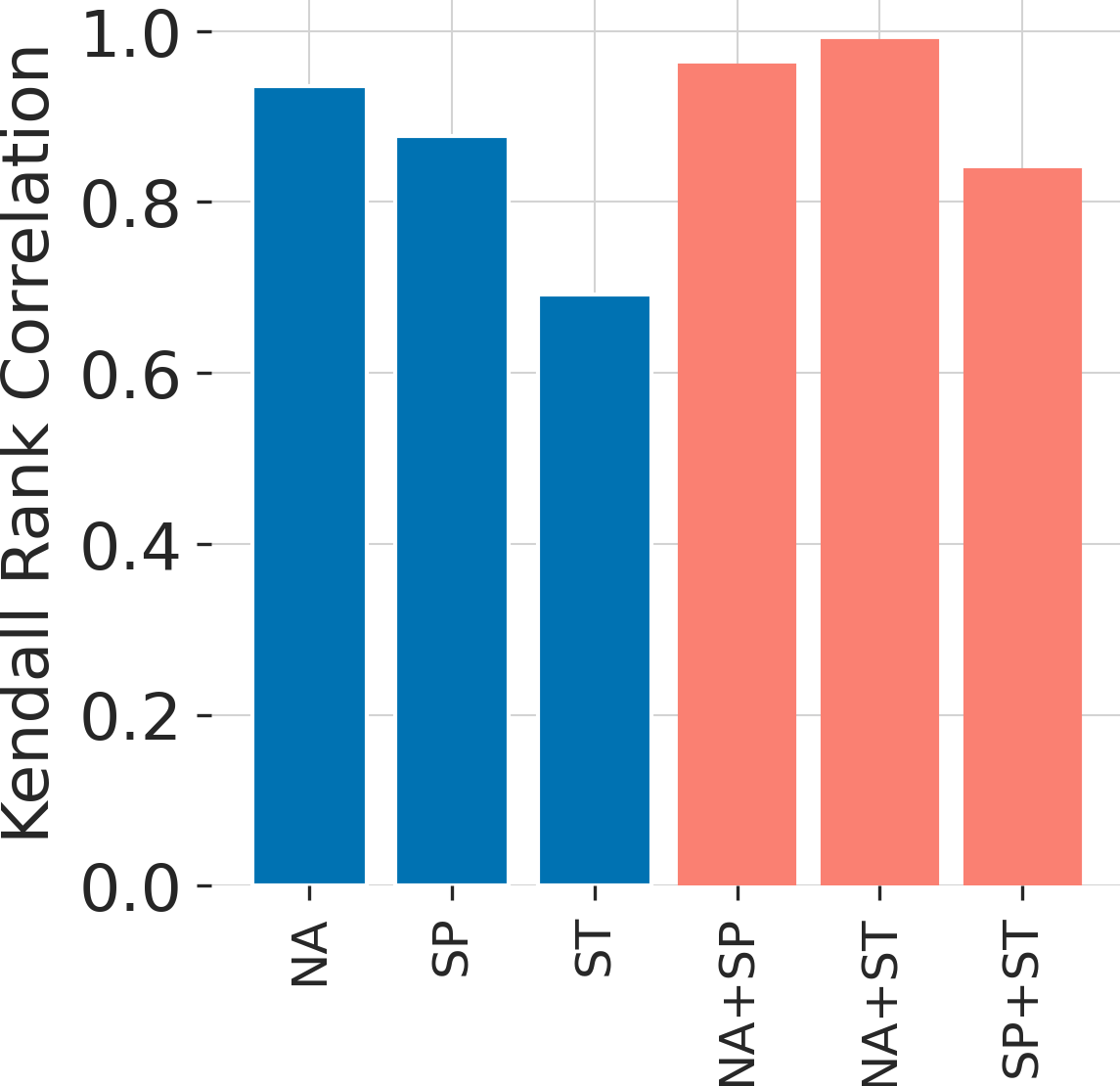}
         \caption{Different categories.}
         \label{fig:VTAB_subcats}
     \end{subfigure}
    \hspace{10pt}
     \begin{subfigure}[b]{0.55\textwidth}
         \centering
         \includegraphics[width=\textwidth]{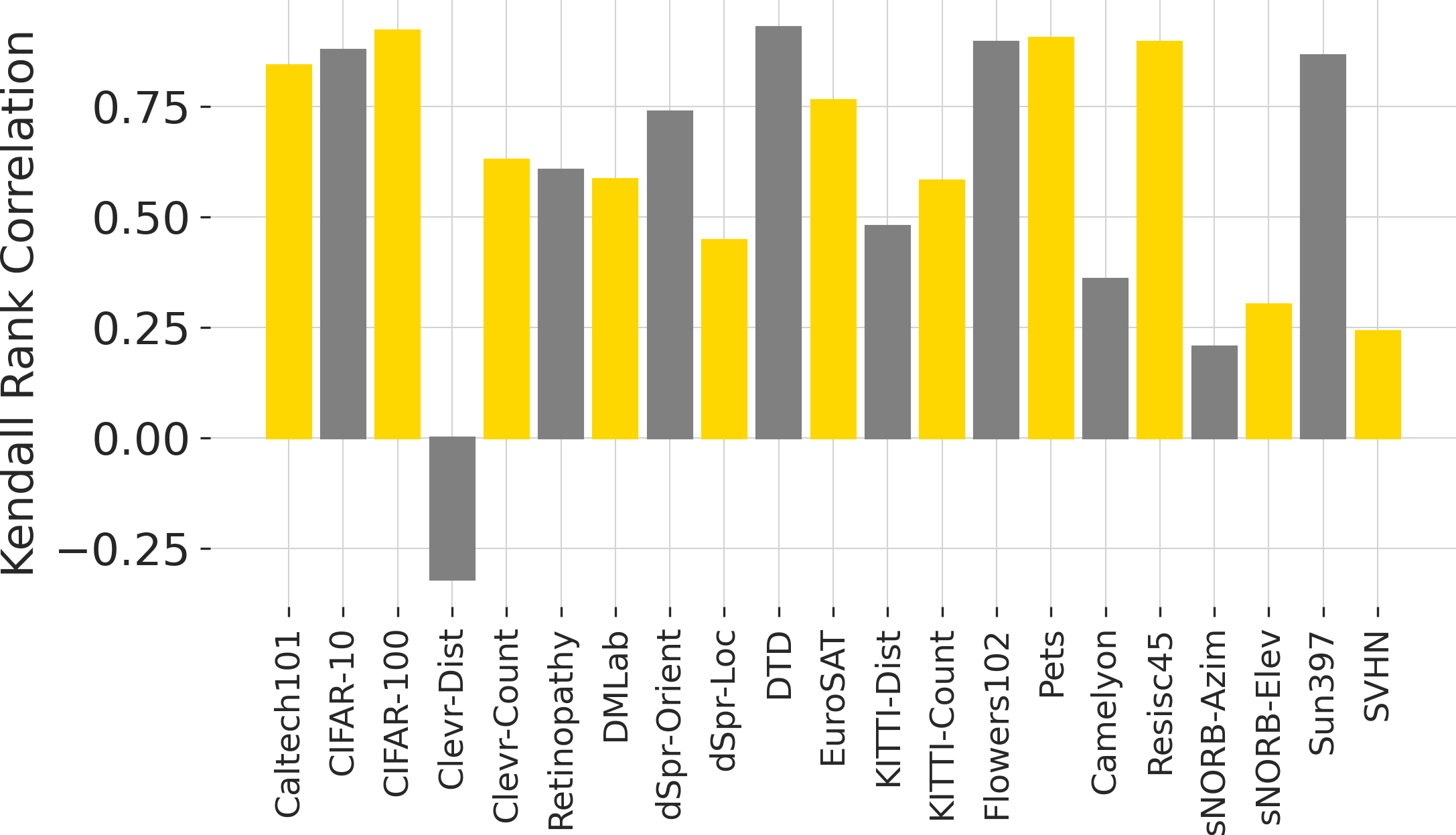}
         \caption{Different task.}
         \label{fig:VTAB_tasks}
     \end{subfigure}
    \caption{Rank correlation between the full VTAB score and the score for subsets of the benchmark.}
    \label{fig:VTAB_corr}
\end{figure}

First, we study the agreement of the aggregated score across all $19$ tasks with the aggregated scores obtained from different combinations of the three task categories: natural (NA), specialized (SP), and structured (ST). Figure~\ref{fig:VTAB_subcats} shows the Kendall rank correlation, when ranking different models based on the full VTAB score and based on the category (combination) score. It can be seen that rankings of models based on different combinations of categories are not always perfectly correlated. For instance, the structured (ST) subcategory has a correlation of $\approx0.7$ with the full VTAB score, thus highlighting rather different aspects of the competing models. A more striking point is the full disagreement of different subcategories on the winning model, i.e. top-1 that is shown in Appendix~\ref{app:vtab_agreement}, where we future present the results that show disagreement in the top-1, 2, and 3 rank positions based on different combinations of sub-categories and tasks. This shows that crowning a model as the winner based on a single score can be suboptimal, and demonstrates how the random nature of task selection can become a lottery that algorithms need to win.

Figure~\ref{fig:VTAB_tasks} also presents the correlations between the rankings based on the individual tasks and the aggregated VTAB score. Unsurprisingly, an even stronger disagreement between rankings is observed (mean Kendall correlation of $\approx0.60$), including tasks with negative correlation.

\subsubsection{Long Range Arena}
\label{sec:lra}
\begin{table}[h]
    \centering
    \small
    \caption{
    Top-$3$ best performing models on the LRA for three cases: the full LRA score, individual tasks and leave-one-out scores. Results on all the possible combinations along with complete model names are reported in Table~\ref{tab:lra-full} in the Appendix~\ref{app:lra_all}.
    }
    \begin{tabular}{l|ccc}
    \hline
    Task & Best Model & Rank-2 & Rank-3 \\
    \hline
LRA Score & B & T & L\\
    \hline
 $t_1$ (Text only)        & L & P & T \\ 
 $t_2$ (Retrieval only) & S & B & L \\
 $t_3$ (ListOps only) & R &  Z & T \\
 $t_4$ (Image only) &  S & P & T \\
 $t_5$ (Path only) & P & L & L \\ 
 \hline
$t_1$ + $t_2$ + $t_3$ + $t_4$ & T  & B & L\\
$t_1$ + $t_3$ + $t_4$ + $t_5$ & B & T & L\\
$t_1$ + $t_2$ + $t_4$ + $t_5$ & S & P & B\\ 
$t_2$ + $t_3$ + $t_4$ + $t_5$ & B & T & L\\
 \hline
    \end{tabular}
    \label{tab:lra}
\end{table}

The Long Range Arena~(LRA; \citet{tay2020long}) is a benchmark designed for aggregated evaluation of long-range Transformer models~\citep{tay2020efficient}. Similar to other benchmark suites, LRA consists of six tasks: ListOps, Long Text Classification, Long Text Retrieval, Pixel-wise Image Classification, and two variants of spatial reasoning based on the path-finder task. The authors rank eleven efficient transformer models by aggregating performance across all six tasks. To demonstrate that here too task selection matters, we computed Top-$3$ rankings of models for each task combination displayed in Table~\ref{tab:lra}. Model name abbreviations are used for brevity and because the actual model names are not important for the purpose of this analysis. Notably, it is easy to see that the identity of each of the top-$3$ changes frequently as the subset of evaluation tasks is changed.

\subsubsection{RL Unplugged}
\begin{figure}[h]
     \centering
     \begin{subfigure}[b]{0.7\textwidth}
         \centering
         \includegraphics[width=\textwidth]{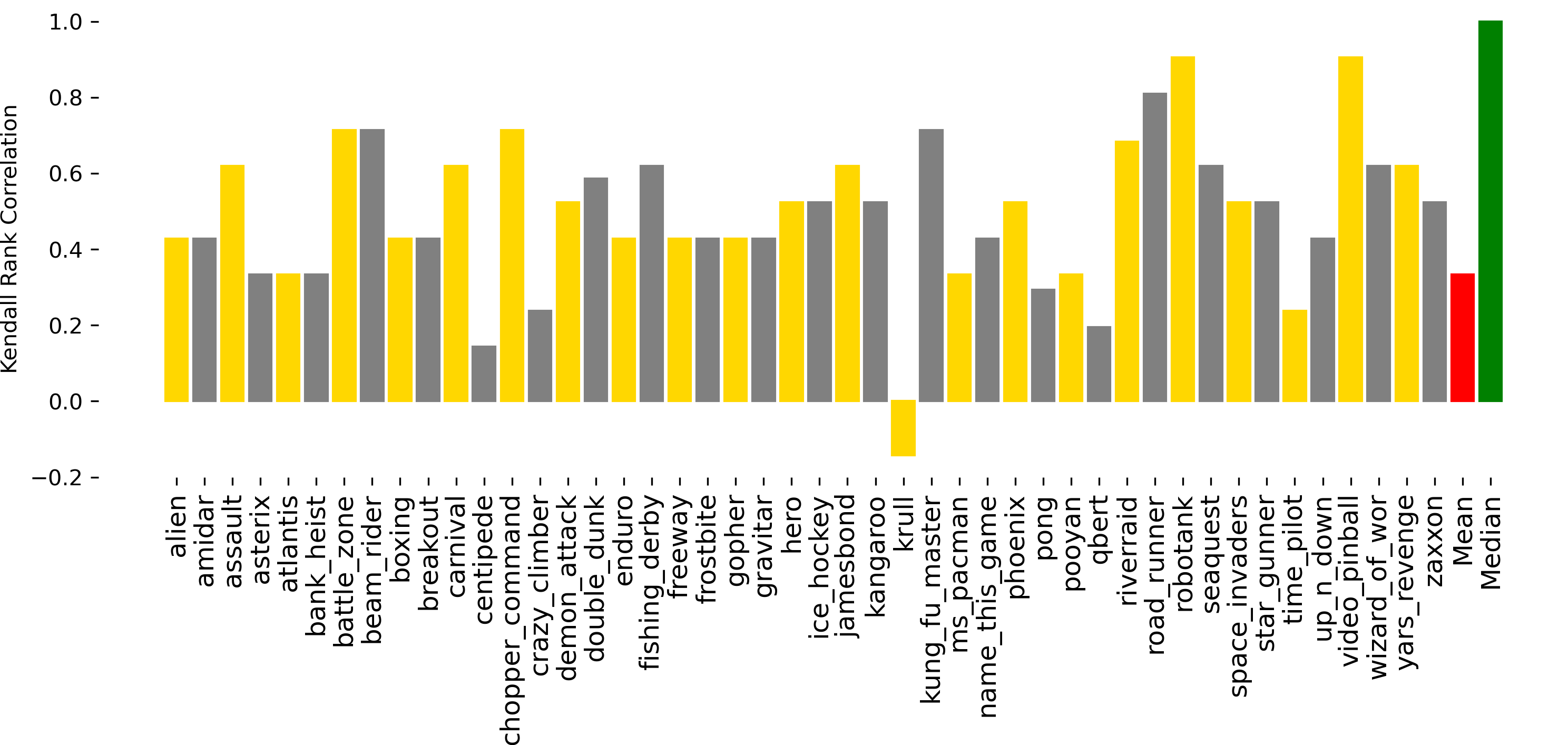}
         \caption{Atari 2600}
         \label{fig:RL_atari}
     \end{subfigure}
    \hspace{10pt}
     \begin{subfigure}[b]{0.25\textwidth}
         \centering
         \includegraphics[width=\textwidth]{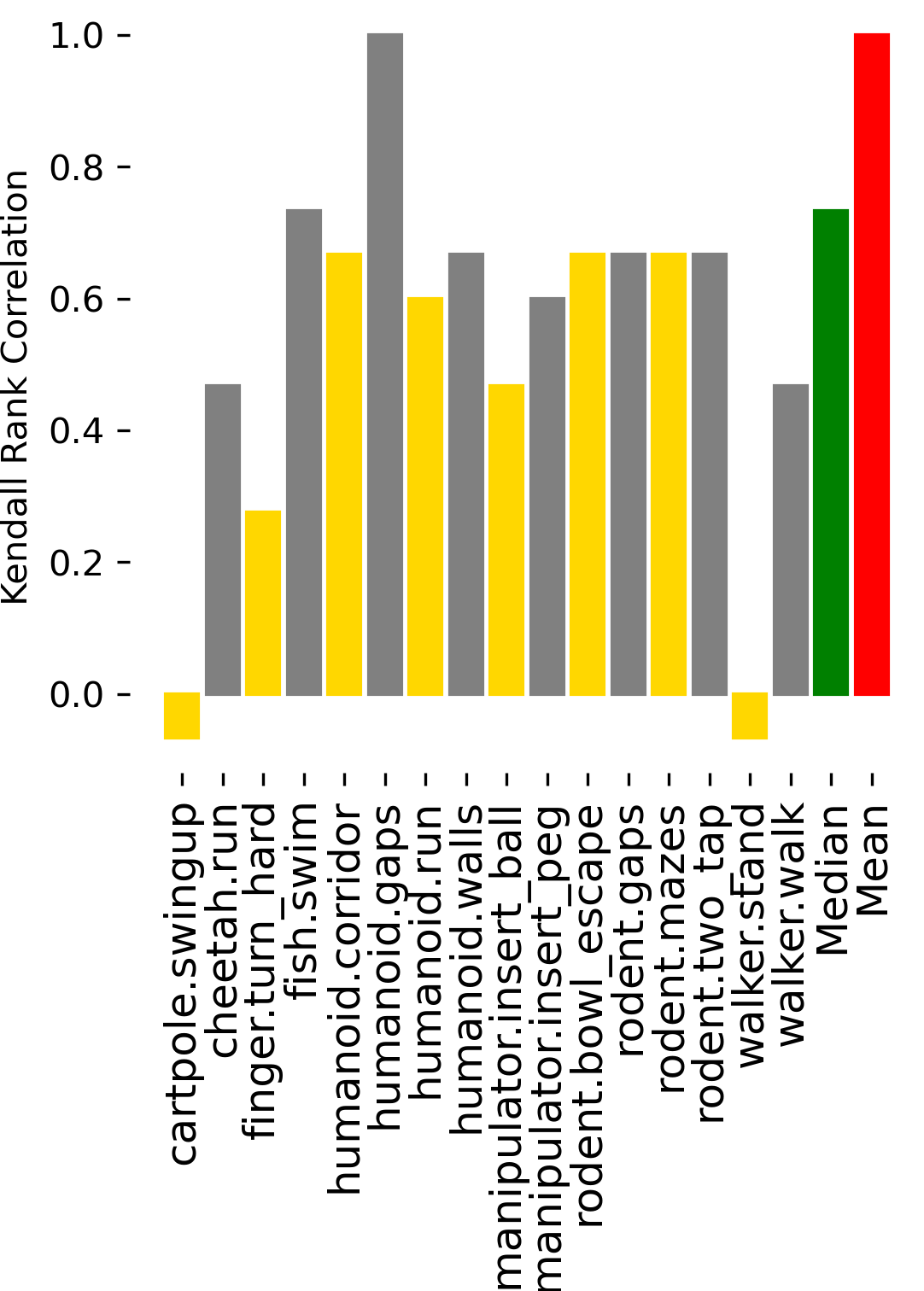}
         \caption{DM Control Suite}
         \label{fig:RL_control}
     \end{subfigure}
    \caption{Rank correlation between the aggregated score and scores from each individual dataset. Note that the common approach in the literature to ingrate scores is ``\emph{median} human normalized performance'' for Atari and ``mean'' for DM controls.}
    \label{fig:RL_corr}
\end{figure}

RL Unplugged~\citep{gulcehre2020rl} is a suite of benchmarks for offline reinforcement learning, where the task for the agent is to learn a policy directly from some logged data that is produced by a system as part of its normal operation, without interacting with the environment at the time of learning. 
In reinforcement learning, in general, it has been shown that varying random seeds alone can lead to a high variance between runs~\citep{henderson2018deep}, and this seed lottery is introducing difficulty in comparing different methods and making conclusions. Here, we study offline RL, where the results are more stable for the sake of focusing a bit more on the task selection bias problem. We will discuss online RL and expand on some other aspects in the context of Section~\ref{section:ALE}.
RL Unplugged introduces a collection of task domains and associated datasets together with a clear evaluation protocol. It includes some widely used domains such as the DM Control Suite~\citep{tassa2018deepmind} and Atari~2600 games~\citep{bellemare2013arcade}, as well as Real-World RL (RWRL) tasks~\citep{dulac2019challenges} and DM Locomotion tasks~\citep{heess2017emergence}.

Here, we study the aggregated scores over multiple tasks in Atari~2600 and DM Control from RL-Unplugged. Atari~2600 consists of $46$ Atari games, and DM Control has $9$ diffident tasks. We use the performance of $7$ differed baselines\footnote{For our analysis we used the data from the ancillary files of ~\citep{schrittwieser2021online}, which can be found in \url{https://arxiv.org/src/2104.06294v1/anc}.} in our analysis.

Figure~\ref{fig:RL_atari} presents the Kendall rank correlation when ranking different models based on their human normalized performance on each task vs the \emph{median} human normalized performance across all tasks. We also show the correlation between median and mean human normalized performance on Atari. Although many papers reported mean performance on Atari as the aggregated score, it is becoming a standard to report median since the mean is potentially less informative, as it is dominated by a few games (e.g. Atlantis) where agents achieve scores orders of magnitude higher than humans do. Figure~\ref{fig:RL_control} also shows the Kendall rank correlation of the mean performance across all tasks with performance on each task as well as the median. 
First of all, in both cases, it can be seen that the ranking of models based on individual tasks can widely disagree the ranking from the aggregated score (average rank correlation in Figure~\ref{fig:RL_atari} is $\approx0.49$ and in Figure~\ref{fig:RL_control} is $\approx0.54$), indicating how solely reporting the aggregated score can send a potentially wrong signal for choosing the best model. 
Moreover, the aggregation strategies, i.e. mean and median in this case do not agree which shows standardizing one over another with the intention of considering only one of them comes at the cost of losing some information.

\subsection{Score and rank aggregation}
So far, we highlighted the issue with reporting a single aggregated score that is supposed to reflect the performance on multiple tasks, by showcasing the disagreement between different subsets of tasks.
One of the main difficulties for aggregating scores of multiple tasks is the lack of a clear mechanism for incorporating the difficulty of tasks into account. This is made more complex by the fact that there are multiple facets to what makes a task difficult. For instance, the size of the training data for different tasks, the number of prediction classes (and consequently the score for a random baseline for the task), distribution shift between the pretraining dataset and the downstream tasks, different performance ranges across tasks, or overrepresenting particular aspects by multiple tasks that introduces biases into averages~\citep{balduzzi2018re}. 
As a concrete example, in the case of VTAB some tasks use the same input data thus upweighting those domains, e.g. CLEVR-Count and CLEVR-Dist use the same data for different tasks, and for this particular example, given the negative correlation between CLEVR-Dist and the mean score, this upweighting effect makes the aggregated score even noisier. 

To address some of these issues, there are alternative ways for ranking models instead of using the mean score across all tasks as the model performance on the benchmark. For instance, One can grouping tasks based on their domain) and use macro-averaging to account for the effect upweighting some domains~\citep{zhai2019large}. Given that using simple averaging for aggregation across multiple tasks, the maximum score is bounded, this may limit the range of performances, implicitly upweighting tasks with more headroom. To address this issue, one can use geometric mean instead of arithmetic mean.
There are also solutions for rank aggregation that ignore absolute score differences in favor of relative ordering~\citep{dwork2001rank, tabrizi2015revisiting}: For instance, the ``average rank'' that is obtained by ranking the methods for each task based on their score and then computing the average ranks across tasks. Another alternatives are, for instance, robust average rank, where, before averaging ranks across tasks, the accuracy is binned into buckets of size 1\% and all methods in the same bucket get the same rank or elimination ranking (which is equivalent to an exhaustive ballot voting system)~\citep{hao2016real}.

\subsection{Human evaluation bias}
Related to the task selection bias we discussed in this section, \emph{human evaluation bias} within a task can also play a role in model selection in some tasks like natural language generation. Lack of consistency in how human evaluation, e.g. due to different levels of expertise, cognitive biases, or even inherent ambiguity in the annotation task can introduce a large variability in model comparisons~\citep{schoch2020problem}. In the context of measuring the reliability in human annotation, it has been shown that selecting a subset of annotators for evaluation may change the performance of models~\citep{van2019best, amidei2018evaluation, schoch2020problem, amidei2020identifying}, which can be framed as ``annotation bias'' that also contribute to the benchmark lottery.

\section{Community bias}
\label{section:community_bias}
Even when viewed as a random process, the task selection bias described in Section~\ref{section:task-selection-bias} alone is sufficient for creating arbitrary selection pressures for machine learning models. We argue however that there is also a higher-level process in which the broader research community influences the task selection, and that counterintuitively leads to the lottery forces not being diminished, but instead more pronounced. This section takes a people perspective of the benchmark lottery and postulates that it is not only the ``gamemasters'' (benchmark proposers) but also the community that contribute to and reinforce it.

While researchers technically have the freedom to select any dataset to showcase their method, this choice is often moderated by the community. In short, the community determines which benchmarks are acceptable for measuring progress on particular problems. 
This process is implicit and not directly observable, as there are no hard rules on permissible benchmarks or tasks.

A common feedback in the review process of scientific publications that any ML researcher will face eventually is a criticism of the choice of benchmark. For example \emph{``the method was not evaluated on X or Y dataset''} or \emph{``the method's performance is not SOTA on dataset Z''}. Over time, ML researchers tend to gravitate to safe choices of tasks and benchmarks. For example, most papers proposing new pretrained language models~\citep{lan2019albert,liu2019roberta,clark2020electra,yang2020xlnet} evaluate on GLUE even if alternatives exist (see example below for further substantiation). In other words, the selection of tasks commonly used in publication is largely driven by the community. Moreover, whether a benchmark is selected as the canonical testbed or not, is not necessarily governed by the quality of the test examples, metrics, evaluation paradigm, or even what the benchmark truly measures.

In fact, an argument that the community is solely responsible for the task selection bias is not without merit, since the community is the final endorser and enforcer of these circumstances. There can be no task selection bias if there is no one to act upon it. To this end, the community might \emph{`double down'} on a benchmark where it becomes almost an unspoken rule for one to evaluate on a particular benchmark. Once a benchmark builds up a following and becomes well-established, it is not hard to imagine that reviewers would ask for results on these benchmarks, potentially regardless of suitability and/or appropriateness. This makes it difficult to fix potentially broken benchmarks.


As foreshadowed, commonly used benchmarks are not immune to containing errors. While these errors are likely to be small (as otherwise they would presumably be noticed early on), they do matter in close calls between competing methods. \citet{northcutt2021pervasive} identified label errors in test sets of $10$ of the most commonly-used computer vision, natural language, and audio datasets; for example, there are label errors in $6\%$ of the examples in the ImageNet validation set. They showed that correcting label errors in these benchmarks changes model ranking, especially for models that had similar performance. In the field of NLP, it was later found in SNLI~\citep{bowman2015large}, which is a dataset for natural language inference (NLI), a large amount of annotation artifacts exists, and it is possible to simply infer the correct label by only using the premise and not the hypothesis~\citep{gururangan-etal-2018-annotation}. It is worth noting that SNLI, being the canonical benchmark for NLI, was easily perceived as mandatory for almost any NLI based research. 

The possibility of having such an issue is not only restricted to the peer review process, but it may extend to the public perception of papers after they are published regardless of whether they went through the peer review process or not. 
The community bias problem can be raised as the community collectively assigning a weighted impact score for doing well on arbitrarily selected tasks. Achieving state of the art on task $Y$ is then deemed significantly less meaningful than doing that for task $X$. Moreover, this is not necessarily done without any explicit reasoning as to why one task is preferred to the other, or even how such a``decision'' was made.

The main concern with respect to the community bias is that research is becoming too incremental and biased toward the common expectations, since a completely new approach will initially have a hard time competing against established and carefully fine-tuned models.

\subsection{Example: GLUE benchmark}
The GLUE benchmark was pitched as a general language understanding benchmark and is an aggregation of 8 datasets that have been previously proposed~\citep{williams2017broad}. We use this as an example of a community bias. To this date, the majority of pretrained LM paper evaluates on the GLUE benchmark. This includes widely recognized and cited papers such as BERT~\citep{devlin2018bert}, ALBERT~\citep{lan2019albert}, RoBERTa~\citep{liu2019roberta}, XLNet~\citep{yang2020xlnet}, ELECTRA, and many others.\footnote{We have manually checked the papers presented ideas to improve pretrained LMs with more than $500$ citations that and in all these papers GLUE has been used for evaluation.}. Given the popularity of GLUE for evaluating these models, it is only imperative that newly proposed models have to shine on GLUE in order to increase their likelihood of acceptance (in the literal sense or by the community). 

Here, it is good to note that seven out of eight tasks in GLUE are actually \emph{matching} tasks that require modeling the relationship between two or more sequences. While it is still unclear how much this problem formulation has to do with natural language understanding, it is clear that this problem formulation favors a certain class of models (e.g., Transformers which has this baked-in cross attention in the encoder). It is easy to see that this conflates an actual advantage in problem formulation (and input setting) with the ability of an encoder model to learn textual representations. While one may argue that a method should reap rewards even for a problem formulation advantage, it is also good to note that many of these cross attention setups are infeasible in practice at scale~\citep{DBLP:journals/corr/abs-1908-10396,seo-etal-2018-phrase}. It is also interesting that, if the tasks in GLUE were swapped for other equally plausible and practical tasks, we might encourage the development of alternative architectures such as pretrained ConvNets in NLP~\citep{tay2021pre}.


\section{Benchmarks are stateful}
\label{section:reusing-benchmark}

With leaderboards and the continuous publication of new methods, it is clear that benchmarks are stateful entities. At any point in time, the attempt of a new idea for beating a particular benchmark depends on the information gathered from previous submissions and publications. This is a natural way of making progress on a given problem. But when viewed from the perspective of the selective pressures it causes, it creates another kind of lottery.

For many machine learning benchmarks, researchers have full access to the holdout set. Although not explicitly, this typically leads to the violation of the most basic datum of ``one should not train on test/holdout set'' by getting inspiration from already published works by others who presumably report only the best of the numerous models they evaluated on the test set.

Beyond that, it is common to copy-paste hyper-parameters, use the same code, and more recently to even start from pre-retrained checkpoints of previous successful models~\footnote{This is in particular common when a paper provides results based on large scale experiments that are not necessarily feasible to redo for many researchers. For instance, the majority of the papers that propose follow up ideas to Vision Transformer~\citep{dosovitskiy2020image} start by initializing weights from the released pretrained models and follow the setups of the original paper. Similarly, several NLP papers use BERT pretrained models and the same hyper-parameters as BERT in their experimental setup.}. In such setups, where the discovery of new models is built on top of thousands of queries, direct or indirect, to the test set, the error rate on test data does not necessarily reflect the true population error~\citep{arora2021rip, blum2015ladder, dwork2015reusable}. 
The adaptive data analysis framework~\citep{dwork2015reusable} provides evaluation mechanisms with guaranteed upper bounds on the difference between average error on the test examples and the expected error on the full distribution (population error rates). Based on this framework, if the test set has size $N$, and the designer of a new model can see the error of the first $i-1$ models on the test set before designing the $i$-th model, one can ensure the accuracy of the $i$-th model on the test set is as high as $\Omega(\sqrt{i/N})$ by using the boosting attack~\citep{blum2015ladder}. In other words, \citet{dwork2015reusable} state that once we have $i\gg N$ the results on the test set are no longer an indication of model quality. 
It has been argued that what matters is not only the number of times that a test set has been accessed as stated by adaptive data analysis, but also how it is accessed. 
Some empirical studies on some popular datasets~\citep{recht2018cifar, yadav2019cold, recht2019imagenet} demonstrated that overfitting to holdout data is less of a concern than reasoning from what has been suggested in~\citep{blum2015ladder}. \citet{roelofs2019meta} also studied the holdout reuse by analyzing data from machine learning competitions on the Kaggle and show no significant adaptive overfitting on the classification competitions. 
Other studies showed that additional factors may prevent adaptive overfitting to happen in practice. For instance,~\citep{feldman2019open, feldman2019advantages} show that in multi-class classification, the large number of classes makes it substantially harder to overfit due to test set reuse.  
In a recent study, \citet{arora2021rip} argue that empirical studies that are based on creating or using new test sets (e.g.~\citep{recht2018cifar, yadav2019cold, recht2019imagenet}), although reassuring in some level, are not always possible especially in datasets concerning rare or one-time phenomena. They emphasize the need for computing an effective upper bound for the difference between the test and population errors. They propose an upper bound using the description length of models that is based on the knowledge available to model designers before and after the creation of a test set. 

All these studies considered, we can conclude that prior information about a model’s performance on the test set may influence future modeling choices and hyperparameter settings, and this adaptive behavior, in principle, may have an effect on generalization and can, to different extents, create a gap between test error and full distribution error. 
Thus, when using a benchmark, we should also think about and clarify answers to several related questions: Do improvements on the benchmark correspond to progress on the \emph{original} problem? How does the number of times, and the way the test (or holdout) set of this benchmark has been accessed relates to the statistical validity of the results reported on it? How far will we get by gaming the benchmark rather than making progress towards solving the original problem?

From the benchmark lottery point of view, the most important aspect of the above phenomena is that the development of new models is shaped by the knowledge of the test errors of all models before it. 
First of all, there had been events in the past where accessing the test set more than others, intentionally, secured a margin for victory in the race~\footnote{\url{https://image-net.org/challenges/LSVRC/announcement-June-2-2015}}. In other words, having the ability to access the test set more than others can be interpreted as buying more lottery tickets. 
Besides, even when there is no explicit intention, the tempting short-term rewards of incremental research polarize people and reinforce the echo chamber effect - leading models are quickly adapted by re-using their code, pre-trained weights, and hyper-parameters are re-used to build something on top of them even faster. Unfortunately, this process makes no time for considering how it affects the statistical validity of results reported on the benchmark. 

Another aspect of benchmarks being stateful is that participating in shared tasks at a later stage is vastly different from the time of its inception. By then the landscape of research with respect to the specific benchmark is filled with tricks, complicated and specialized strategies, and know-how for obtaining top performance on the task. The adapted recipes for scoring high are not necessarily universal and may be applicable only to a single narrow task or setup. For example, a publication might discover that a niche twist to the loss function produces substantially better results on the task. It is common for all papers subsequently to follow suit. As an example, the community realized that pre-training on MNLI is necessary for obtaining strong performance on RTE and STS datasets~\citep{liu2019roberta,clark2020electra}, and this became common practice later on. Experience shows that it is not uncommon for benchmark tasks to accumulate lists of best practices and tricks that are dataset- and task-specific~\footnote{As an example, for achieving scores that are comparable to top-ranked models on the GLUE benchmark, there are a series of extremely specific actions and setups used in pretraining/finetuning that are known as ``standard GLUE tricks'' introduced/used by submissions to the leaderboard~\citep{liu2019roberta, yang2019xlnet, lan2019albert}. Check the Pre-training and fine-tuning details in the appendix of~\citep{clark2020electra}.}. Whether a novel algorithm is able to make use of these tricks (or whether they are available at all) is again a form of lottery, in which models that cannot incorporate \emph{any} of the earlier tricks are significantly disadvantaged.

\section{Rigging the lottery: making your own luck}
\label{section:rigging-the-lottery}
For some tasks and problems, there are already standard benchmarks and established setups that are followed by most of the community. However, for some others, inconsistencies in the employed benchmarks or reported metrics can be observed. This diversity of evaluation paradigms makes comparisons between publications extremely difficult.
Alternatively, in some cases, there is simply no standard benchmark or setup, either because the problem is still young, or because there has never been an effort to unify the evaluation. Sometimes this is due to the high computational cost of proper evaluation, like when reporting variance over multiple random seeds is important~\citep{bouthillier2019unreproducible}. While in other instances, the root cause is of behavioral nature, where researchers prefer to showcase only what their method shines at - oftentimes to avoid negative reviews, unsuccessful experiments, although performed, are simply not reported. Here, we study two known examples of this issue, which we refer to as \emph{rigging the lottery}.


\subsection{Recommender systems and benchmark inconsistencies}
The goal of recommender systems (RecSys) research is to build systems that understand users' interests and behaviors towards a list of potential items to recommend. Deep learning-based recommender systems are trained on large quantities of implicit user feedback, such as users clicking on or otherwise interacting with items~\citep{zhang2019deep}.


Unlike the fields of NLP or CV, there are no well-established evaluation setups for recommender systems that provide canonical ranked lists of model performance. While there has been a famous Netflix prize challenge$\footnote{\url{https://netflixprize.com/index.html}}$, this dataset has not been extensively used in academic research or for benchmarking new models. Moreover, even popular datasets like MovieLens~\citep{10.1145/2827872} or Amazon Reviews~\citep{he2016ups} generally do not have a canonical test split, metric or evaluation method. Therefore, it is still quite unclear about which modern RecSys method one should adopt, as model comparisons are difficult to interpret~\citep{dacrema2019we}.

Furthermore, RecSys evaluation is also very challenging for a number of reasons. (\emph{i}) Different recommendation platforms tackle slightly different problems (e.g retrieval~\citep{yi2019sampling}, ranking (\citep{pei2019personalized}), or multitask learning (\citep{zhao2019recommending})), and each requires their own evaluation setup. (\emph{ii}) As is common for user interacting systems, user's reaction towards different algorithms can be different. Constructing offline datasets of user behaviors from an existing system creates an off-policy evaluating challenge~\citep{swaminathan2016off}. (\emph{iii}) A real-world recommendation system trains on billions of users and items, the scale of user-item interactions makes it extremely difficult to create a complete dataset containing all possible user-item interactions~\citep{he2016fast}. As a result, evaluation setups in many recommender system papers tend to be arbitrary.

There exists a small number of public datasets (see Appendix~\ref{app:rec_sys}), such as MovieLens~\citep{10.1145/2827872} or Amazon Product Review~\citep{he2016ups} that are commonly used for evaluating recommender systems. However, even these datasets are tweaked differently in various publications, leading sometimes to contradictory results~\citep{rendle2020neural,zheng2019mars}. For example, some papers use Hit Ratio and NDCG as evaluation metrics~\citep{he2017neural}, while others resort to using Recall@K~\citep{zheng2019mars}. Interestingly, in this particular example, the same methods reverse their performance when a different metric is used. Holdout test sets can also be created differently, with some papers for example using random split~\citep{beutel2017beyond} and others using an out-of-time split~\citep{zhang2020model}.

While the majority of this paper discusses cases where a standardized benchmark may lead to biased progress in the ML community, here we instead discuss the \emph{exact opposite} - implications of having no consensus datasets or evaluation setups. Having no unified benchmark for the community to make progress on has numerous flaws. To name a few, (\emph{i}) this hinders progress in the field, while possibly (\emph{ii}) creating an illusion of progress. It is not surprising that under these circumstances researchers (potentially unknowingly) tend to find good experimental setups that fit their models instead of the other way around. 


\subsection{ALE and evaluation setup inconsistencies}
\label{section:ALE}
An example of a benchmark that hundreds of papers have used as a testbed, while simultaneously employing a number of distinct experimental evaluation protocols is the Arcade Learning Environment (ALE) which is based on Atari~2600 games~\citep{mnih2013playing}. The main aspects in which evaluation setups in different papers using ALE diverge are different metrics used for summarizing agent performance, and the different mechanisms used for injecting stochasticity in the environment~\citep{machado2018revisiting}.

For example, different assumptions can be made for determining episode termination. While in some publications episodes terminate when the game is over~\citep{bellemare2013arcade,hausknecht2014neuroevolution, liang2015state,lipovetzky2015classical, martin2017count}, while others papers choose to terminate the training episodes for a subset of the games when the agent loses a life~\citep{mnih2016asynchronous, nair2015massively, wang2016dueling, van2016deep}.

Another major disagreement in evaluation strategies for ALE, also comes from using different parameters used for the evaluation setup. For example, some papers use a non-default value for the skipframe parameter\footnote{When predicting the action given the state, it is often done for every $k$-th frame, where $k$ is the skipframe hyper-parameter.} in their baseline models~\citep{mnih2015human}. 
Alternatively in some publication, methods are evaluated for each $2\times10^5$ frames~\citep{pritzel2017neural}, while in others methods are evaluated every $10^6$ frames~\citep{mnih2013playing, mnih2016asynchronous}.
Another observation is the difference between the number of games used in the evaluation setups. For instance, \citet{mnih2015human} use $49$ games, while \citet{van2016deep, wang2016dueling} use $57$.
Moreover, for the hyper-parameter tuning, sometimes papers use the entire suite of games as the validation set~\citep{bellemare2013arcade}, while in other cases hyperparameters are optimized on a per-game basis~\citep{jaderberg2016reinforcement}.

Yet another inconsistency is in reporting the results in terms of the variety of different summary statistics used to describe them, which makes direct comparisons between ideas difficult~\citep{machado2018revisiting}. To make matters worse sometimes sufficient statistics to make a judgment on the quality of the models are not provided. As an example, in~\citep{bellemare2013arcade}, the main results are reported as the average performance of the method as well as the best run without mentioning the variance or the standard error of the mean. This is particularly problematic for reinforcement learning, where it has been shown that often the variance between runs can be so large as to create statistically different distributions just by varying random seeds~\citep{henderson2018deep}.

The final contentious aspect of ALE that we highlight is the way that various publications choose to inject stochasticity into the environment. ALE is fully deterministic, thus it is possible to get good scores by simply memorizing the ``right'' action sequence, rather than learning to make good decisions in a variety of game scenarios (i.e. learning an open-loop policy). With this in mind, to encourage and evaluate agent robustness, various ideas were developed to add forms of stochasticity to ALE~\citep{bellemare2013arcade}. Unfortunately, these methods are not necessarily consistent with each other.



\section{What can we do?}
\label{section:what_can_we_do}
While the previous sections of the paper focused on the challenges that arise from the lottery-like interaction between ML benchmarks and the research community, here we would like to show that there are reasons to be optimistic about future developments in this regard. We present suggestions for improving the idea benchmarking process in ways that make it less of a lottery.
These recommendations can be also framed as checklists\footnote{Similar to the reproducibility checklist \url{https://www.cs.mcgill.ca/~jpineau/ReproducibilityChecklist.pdf}~\citep{dodge2019show}} for different parts of the process, like making benchmarks, using benchmarks, evaluation of a new ideas. Appendix~\ref{app:checklist} presents a proposed benchmarking checklist for the review process.

\subsection{Investing in making guidelines}
Assessing the quality of ML algorithms is a complex task, not the least due to the diversity of aspects and contexts in which algorithms are compared (e.g. based on performance, theoretical computational efficiency, efficiency on specific hardware, fairness w.r.t. selected attributes, etc). This adds an extra dimension of complexity to the already difficult task of setting up a benchmark.

Given the pitfalls discussed in this paper, it is not impossible to imagine a benchmark being set up that inadvertently misrepresents the progress and causes an ill-fated shift in research questions and priorities for (a part of) the research community. The risks of this happening can be minimized by standardizing the recipe for creating new benchmarks. We believe that the development of shared standards and guidelines for future benchmarks can be the first step towards solving the ``rigging the lottery'' problem described in Section~\ref{section:rigging-the-lottery}. We discuss such guidelines in the remainder of this section. 

\paragraph{Benchmark building blocks} The ultimate goal of a benchmark is to provide an opportunity to learn more about a specific problem, which is achieved by quantifying progress made on that problem. So \emph{the problem} or the task itself has a central role when designing a new benchmark. Underinvestment in a clear definition of the problem can offset the outcome of the comparisons, while having a clear set of objectives can assist in identifying the right datasets and usage guidelines for the benchmark.

\emph{Datasets} form the core of the benchmark, and great care should be taken to collect or select data for the benchmark that appropriately captures the problem of interest, has the ``right'' dataset size, and faithfully captures the distribution of the larger population.

Finally, quantifiable progress necessitates the use of shared \emph{metrics}, which similarly are an essential part of a benchmark. Their choice requires great care, as at the end of the day, metrics reflect the progress and dictate future research directions.

The gola, datasets, and metrics form the essential building blocks of a benchmark (suite),  when chosen without deliberation, can respectively feed into the community (Section~\ref{section:community_bias}) and task selection biases (Section~\ref{section:task-selection-bias}) and rigging the lottery (Section~\ref{section:rigging-the-lottery}) behaviours.

\paragraph{Guidelines for creating benchmarks} Given this, we believe that investing more into shared guidelines for creating new benchmarks can be extremely beneficial to the long-term health of the research community. In our view, such guidelines should include the current best practices and aspects that require special attention; and should highlight potential concerns for issues that may emerge in the future when different models and algorithms are applied to the benchmarks. 

Fortunately, there have been some efforts in providing guidelines and best practices for making new benchmarks. For example, \citet{covariant-robotics-benchmark} discusses the need for how robotic warehouse picking benchmarks should be designed to assess the ability of robots in terms of their out of the box performance (i.e. their success rate the first time a new item or unknown scene is encountered), learning speed (i.e. the time required to adapt to new experiences, objects or scenarios) and learning potential (i.e. ability to master new experiences), and provides tips on object selection, scene design, execution, and type of analysis that should be done for proper evaluation. \citet{kiela2021dynabench} proposed a framework for benchmarking in NLP that unifies dataset creation, model development, and model assessment, in a dynamic way with humans and models in the loop. Although their proposal aims at addressing issues with traditional static benchmarks, it also sets clear standards for making new tasks and benchmarks.
\citet{denton2020bringing} look at the dataset construction process with respect to the concerns along the ethical and political dimensions of what has been taken for granted, and discuss how thinking about data within a dataset must be holistic, future-looking, and aligned with ethical principles and values. \citet{bender2018data} also proposed using data statements for NLP datasets in order to provide context that allows users to better understand how experimental results on that dataset might generalize, how software might be appropriately deployed, and what biases might be reflected in systems built on the software.

Another issue that is rather common in many of the ML benchmarks is that the top-scoring models are often overly complex or specialized. Sometimes they are ensembles of different methods or there are fancy techniques involved in the training process. 
Sometimes these complex methods grand-father the benchmark by staying at the top position for a long time, which might kill the potential of elegant models having their moment of fame and get the chance of shining and being picked by the researchers as good starting points.
One idea to address this issue can be rewarding simplicity and ease of adaptation, and different benchmarks and tasks, there may be ways to define a quantitative metric to be taken into account when ranking models. 
From a more general point of view, one of the main issues here is the assumption that there is a single ideal model, while the appropriate model is, in fact, a function of context. In other words, dimensions considered for ranking models are inherently dynamic and can change as the application changes. Thus besides simplicity, many factors can play a role in choosing the best model, given the requirements of the context.


In Section~\ref{section:rigging-the-lottery} we pointed out that sometimes the blocking factor or conduction rigorous evaluation is the high computational costs, in particular for the academic environment. For instance, analyzing all possible sources of variance in the performance is prohibitively expensive. As a potential solution to this problem, the community can invest more in setting up initiatives like reproducibility challenges\footnote{For instance \url{https://paperswithcode.com/rc2020},  \url{https://reproducibility-challenge.github.io/iclr_2019/}, or \url{https://reproducibility-challenge.github.io/neurips2019/}} or specialized tracks at conferences that offer help in terms of expertise, infrastructure, and computational resources for extensive evaluation to the papers submitted to that conference.

\paragraph{Guidelines for benchmark usage} Besides the necessity of making guidelines for ``how to make new benchmark'', it is important to have clear guidelines for ``how to use a benchmark'', which for instance includes the exact setup that the benchmark should be used for evaluation or how the results should be reported. This would be a great help with reducing the instances of rigging the lottery prevalent in some domains (Section~\ref{section:rigging-the-lottery}).

Most modern benchmarks have a ``how to'' section that focuses on the technical aspects, such as loading the data, or installing helper libraries, and using utility functions, running the baselines or the evaluation scripts, or submitting entries to the leaderboard. This is very helpful with getting started with a new benchmark, but is insufficient to prevent inconsistencies between publications in the adapted setups or the way results are presented. Such inconsistencies make it difficult to understand the progress in the field based on these benchmarks. We believe that these ``how to'' sections can be expanded into guidelines providing a set of recommendations for a process to follow when using the benchmark in order to avoid shortcuts that undermine the validity of findings. Furthermore, because in some cases, like those that the method is supposed to work with open data, the difficulty of using the default benchmark setup is the main cause for researchers to diverge from that setup, the guidelines could also focus on establishing standard methodologies for principled comparison and analysis of experimental results (see also Section~\ref{section:statistical-significance}).

Top scoring approaches on any benchmark are often overly complex, overly-specialized models, heavy multi-scale ensembles, fancy training techniques, etc. How do we organize benchmarks/metrics for the "simplest baseline that just works".

There are also several efforts targeting this goal. For instance, \citet{albrecht2015reports, machado2018revisiting} propose specific standards for the ALE benchmark (discussed in Section~\ref{section:ALE}) to have more comparable numbers across different baselines. There is also literature highlighting the issue of reporting a single number as the quality of a model, which encourages blindly optimizing for climbing a leaderboard without taking other aspects into account. \citet{ethayarajh2020utility} argue against ranking models merely based on their performance and discuss the example of a highly inefficient model that provides little utility to practitioners but ranks well on leaderboards. To shed light on such instances, they propose to always report \emph{model size}, \emph{energy efficiency}, \emph{inference latency}, and metrics indicating model \emph{robustness} and \emph{generalization to the out-of-distribution data}.
\citet{gebru2018datasheets} proposed that every dataset be accompanied by a datasheet that documents its motivation, composition, collection process, recommended uses, etc with the goal of increasing transparency and accountability, mitigating unwanted biases in ML systems, facilitating greater
reproducibility, and helping researchers and practitioners select more appropriate datasets for their chosen tasks.

Another important problem that can benefit from established regulation is the hyper-parameter tuning budget used by researchers to improve their model performance. Spending enough time and compute to precisely tune hyper-parameters of the model or the training process can improve the results a great deal~\citep{li2018system, bello2021revisiting, steiner2021train}.
Given that, a guideline on limiting the budget for the hyper-parameter tuning can curb the improvements that are solely based on exhausting hyper-parameter search and gives a chance to have comparisons that are tied less to the computational budget of the proposing entity, but more to the merits of the methods themselves.

\paragraph{Guidelines for conferences and reviewers}
There have been attempts to ameliorate the problems related to the benchmark lottery, especially its community biases and the statefulness aspects (Sections~\ref{section:community_bias} and \ref{section:reusing-benchmark}). For example, NLP conferences have specially called out \emph{``not being SOTA''} as an invalid basis for paper rejection\footnote{\url{https://2020.emnlp.org/blog/2020-05-17-write-good-reviews}.} We believe it is possible to leverage education through the review process in order to alleviate many negative aspects of benchmark lottery.

As an example, we can make sure that in the review process, scores on a particular benchmark are not used for immediate comparison with the top-ranking method on that benchmark, but rather as a sanity check for new models and simply an efficient way of comparing against multiple baselines. This way, fundamentally new approaches will have a chance to develop and mature instead of being forced to compete for top performance right away or get rejected if not succeeded in the early attempts.

\subsection{Statistical significance testing}
\label{section:statistical-significance}
The presence of established benchmarks and metrics alone does not necessarily lead to a steady improvement of research ideas; it should also be accompanied by rigorous procedures for comparing these ideas on the said benchmarks. 
For example, \citet{armstrong2009improvements} discuss the importance of comparing improvements to the strongest available baselines, and how the lack of such practice was responsible for a decade of stagnation experienced by the field of information retrieval in the early 2000s. Luckily modern day ML research largely overcame this issue by setting up strong baseline expectations and incentives for making research easily reproducible (e.g. citations and general interest in publications sharing source code). However, the question of how do we know that if a new model $B$ is \emph{significantly} better than its predecessor model $A$ remains anything but solved 10 years later~\citep{lin2021significant}.

\paragraph{Benchmark results as random samples} Machine learning models are usually trained on a training set and evaluated on the corresponding held out test set, where some performance metric $m$ is computed. Because model training is subject to sources of uncontrolled variance, the resulting metric $m$ should be viewed as a single sample from the distribution describing the model's performance. Because of that, deciding which of the two models is better based on point estimates of their performances $m_A$ and $m_B$ may be unreliable due to chance alone. Instead distributions of these metrics $p(m_A)$ and $p(m_B)$ can be compared using statistical significance testing to determine whether the chance that model $A$ is at least as good as model $B$ is low, i.e. $p(A\leq B)<\alpha$ for some \emph{a priori} chosen significance level $\alpha$. Estimation of $p(A\leq B)$ forms for the crux of statistical significance testing. It can be done either by using parametric tests that make assumptions on distributions $p(m_A)$ and $p(m_B)$ and thus often need fewer samples from these distributions, or by using non-parametric tests that rely on directly estimating the metric distributions and require more samples. 

The popularity of standardized benchmarks and exponential growth in the amount of research that the ML community has experienced in recent years\footnote{\url{https://neuripsconf.medium.com/what-we-learned-from-neurips-2020-reviewing-process-e24549eea38f}} exacerbate the risk of inadvertently misguiding research through lax standards on declaring a model as an improvement on the SOTA. Indeed, if point estimates are used in place of statistical significance testing procedures, sampling $m'_A \sim p(m_A)$ and $m'_B \sim p(m_B)$ such that $m'_B > m'_A$ is only a matter of time, even if performance of the two models is not actually different. Note that this is \emph{not} the same as the issue described in Section~\ref{section:reusing-benchmark}, but could instead be thought of as winning a lottery if you purchase enough lottery tickets.

\paragraph{Beyond a single train-test split}
Unfortunately, researchers rarely go through the process of collecting strong empirical evidence that model $B$ significantly outperforms model $A$. This is not surprising. As discussed in \citet{Bouthillier2021}, obtaining such evidence amounts to running multiple trials of hyper-parameter optimization over sources of variation such as dataset splits, data ordering, data augmentation, stochastic regularisation (e.g. dropout), and random initialization to understand the models' variance, and is prohibitively expensive\footnote{Although \citet{Bouthillier2021} also propose a pragmatic alternative to the exhaustive study of all source of variation.}. If studied at all, mean model performance across several random parameter initializations is used for declaring that the proposed model is a significant improvement. This is vastly sub-optimal because dataset split contributes the most to model variance compared to other sources of variation~\citep{Bouthillier2021}. However, providing multiple dataset splits to estimate this variance is not standard practice in benchmark design.

Benchmarks typically come with a single fixed test set, and thus could even be said to unintentionally discourage the use of accurate statistical testing procedures. This is particularly problematic for mature benchmarks (i.e. Stage~4; see Section~\ref{section:life-of-a-benchmark}), where the magnitude of model improvements may become comparable to the model variance. Systematic variance underestimation may lead to a series of false positives (i.e. incorrectly declaring a model to be a significant improvement) that stall research progress, or worse - lead the research community astray by innovating on ``improved overfitting'' in place of algorithmic improvements. Going forward, one way of addressing this limitation is to design benchmarks with \emph{multiple} fixed dataset splits. As an added benefit, model performance reported across such standardized splits would also enable the application of a variety of statistical tests not only within the same study, but also across publications.

Contrary to the above concern, a recent study of the progress on the ImageNet benchmark~\citep{recht2019imagenet, deng2009imagenet} found that in image classification overfitting on the test does not appear to be an issue despite the benchmark's popularity. \citeauthor{recht2019imagenet} found that algorithms ranked consistently when re-trained on a different train-test split, and that observed drops in performance on the new test split could be explained by the ``difficulty'' of the test images rather than overfitting on the original test split. It may thus also be acceptable to take a shortcut and rely on a single train-test split for benchmarks so long as there are also periodic studies such as~\citep{recht2019imagenet} that double check that the research community is still making progress. Having said that, the decision of whether such shortcut can be used should be made on a case-by-case basis, and there is no harm in producing multiple standardized train-test splits for a new benchmark even if a single split ends up being used.

\paragraph{Benchmark design with statistical testing in mind} The choice of a suitable statistical testing procedure is non-trivial. It must consider the distribution of the metric $m$ that is being compared, the assumption that can be safely made about the distribution (i.e whether a parametric test is applicable or a non-parametric test should be used), the number of statistical tests performed (i.e. whether multiple testing correction is employed) and can also change as the understanding of the metric evolves~\citep{Demvsar2006,Bouthillier2021,lin2021significant}. We, therefore, recommend that benchmark design is accompanied by the recommendation of the suitable statistical testing procedures, including the number dataset splits discussed above, number of replicates experiments, known sources of variance that should be randomized, the statistic to be computed across these experiments and the significance level that should be used for determining statistically significant results. This would not only help the adoption of statistical testing for ML benchmarks, but also serve as a centralized source for best practices that are allowed to evolve. A detailed discussion of statistical testing is outside of the scope of this paper, and we refer interested readers to~\citep{Bouthillier2021, Dror2017} for an overview of statistical testing procedures for ML.

\paragraph{Beyond a single dataset}
Often we are interested in understanding whether model $B$ is significantly better than model $A$ \emph{across a range of tasks}. These kinds of comparisons are facilitated by benchmarks that span multiple datasets (e.g. VTAB or GLUE). Already the question of what it means to do better on a multi-task benchmark is non-trivial due to the task selection bias (see Section~\ref{section:task-selection-bias}) - is it sufficient for model $B$ to do better on average; or should it outperform model $A$ on all tasks? It is not surprising that the statistical testing procedures for such benchmarks are also more nuanced - the answer to this question leads to different procedures. It is unclear whether the average metric across datasets, a popular choice for reporting model performance, is meaningful\footnote{In fact for that reason it was not a popular choice until recently~\citep{Demvsar2006}.} because the errors on different datasets may not be commensurable, and because models can have vastly different performance and variances across these datasets. For this reason, more elaborate procedures are required. For example, for the case when we are interested in seeing whether $B$ outperforms $A$ on average \citet{Demvsar2006} propose to ignore the variance on individual datasets and treat the model $A$~and~$B$'s performance across datasets as samples from two distributions that should be compared. They recommend that the Wilcoxon signed-rank should be used in such a setup; but the recommended can have limited statistical power when the number of datasets in the benchmark is small. Alternatively, for cases when we are interested in seeing whether $B$ is better than $A$ on all datasets \citet{Dror2017} propose to perform statistical testing on each of the datasets separately while performing multiple testing corrections. Here again the ``right'' statistical testing procedure depends on the benchmark, its composition, and the criteria for preferring one model over another; and we believe that the community would benefit if these questions were explicitly answered during benchmark design.

\subsection{Rise of living benchmarks}
Another major issue for many popular benchmarks is ``creeping overfitting'', as algorithms over time become too adapted to the dataset, essentially memorizing all its idiosyncrasies, and losing the ability to generalize. This is essentially related to the statefulness of benchmarks discussed in Section~\ref{section:reusing-benchmark}. Besides that, measuring progress can be sometimes chasing a moving target since the meaning of progress might change as the research landscape evolves.

This problem can be greatly alleviated by for instance changing the dataset that is used for evaluation regularly, as it is done by many annual competitions or reoccurring evaluation venues, like WMT\footnote{\url{http://statmt.org/}} or TREC\footnote{\url{https://trec.nist.gov/}}.  Besides that, withholding the test set and limiting the number of times a method can query the test set for evaluation on it can also potentially reduce the effect of adaptive overfitting and benchmark reuse.
In a more general term, an effective approach is to turn our benchmarks into ``living entities''. If a benchmark constantly evolves, for instance, adds new examples, adds new tasks, deprecates older data, and fixes labeling mistakes, it is less prone to ``tricks'' and highly robust models would find themselves consistently doing well across versions of the benchmark. As examples of a benchmark with such a dynamic nature, GEM is a living benchmark for natural language generation~\citep{gehrmann2021gem} or Dynabench~\citep{kiela2021dynabench} proposes putting humans and models in the data collection loop where we continuously reevaluate the problem that we really care about.

\section{Epilogue}
Ubiquitous access to benchmarks and datasets has been responsible for much of the
recent progress in machine learning. We are observing the constant emergence of new benchmarks. And on the one hand, the development of benchmarks is perhaps a sign of continued progress, but on the other hand, there is a danger of getting stuck in a vicious cycle of investing in making static benchmarks that soon will be rejected due to the inflexible flaws in their setup, or lack of generality and possibility for expansion and improvements. 
We are in the midst of a data revolution and have an opportunity to make faster progress towards the grand goals of artificial intelligence if we understand the pitfalls of the current state of benchmarking in machine learning. 
The ``benchmark lottery'' provides just one of the narratives of struggling against benchmark-induced model selection bias. 
Several topics we touched upon in this paper are discussed in the form of opinions or with a minimum depth as a call for further discussion. We believe each subtopic deserves a dedicated study, like how to better integrate checks for ethical concerns in the mainstream evaluation of every existing benchmark, how to develop tools and libraries that facilitate the rigorous testing of the claimed improvements, or a deep investigation of the social dynamics of the review process and how to improve it.  
In the end, there are many reasons to be excited about the future - the community is continuously taking positive delta changes that contribute to fixing issues with measuring progress in the empirical machine learning.

\section*{Acknowledgement}
We would like to thank Lucas Beyer, Alexander Kolesnikov, and Xiaohua Zhai for their help and the information they provided us to analyze the VTAB benchmark. Also Lasse Espeholt and Julian Schrittwieser for the pointers they provided for the RL Unplugged analyses and finally Sharan Narang for helping us with the data needed for the SuperGLUE analyses. We are also grateful to Samira Abnar, Jakob Uszkoreit, Avital Oliver, and Anurag Arnab for their fruitful comments and inspiration on the topic along the way. We would like to thank Sara Hooker, Jason Baldridge, Alex Hanna, Emily Dentone, and Douglas Eck for the feedback they provided to improve the paper. And finally, we would like to also thank anonymous reviewers of the datasets and benchmarks track of NeurIPS for their feedback, fresh ideas, and suggestions to improve the paper.

\vskip 0.2in
\bibliographystyle{plainnat}
\bibliography{references}

\newpage
\input{appendix}

\end{document}

%% file: appendix.tex
\section{Appendix}
\appendix

\section{SuperGLUE: Ranking of models on different combinations of tasks}
\label{app:superglue_comb}
\begin{figure}[h]
    \centering
    \includegraphics[width=0.72\textwidth]{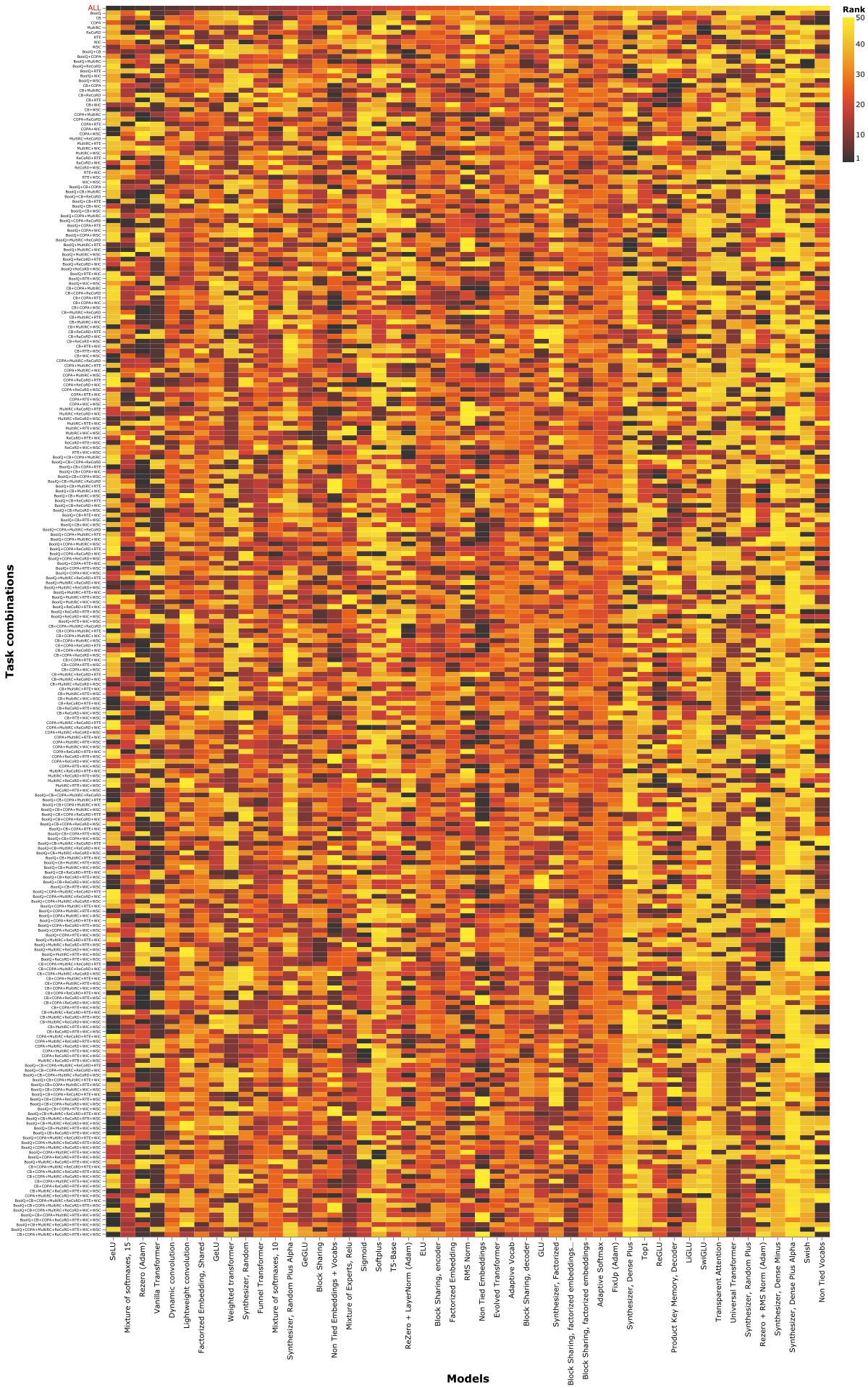}
    \caption{\small{Performance of different models on different combinations of tasks in in terms of their rank on the SuperGLUE benchmark. Models are sorted on the $x$-axis based on their rank when evaluated on ``ALL'' tasks. We can observe that there no clear and strong pattern or correlation in different combinations of tasks compared to the full benchmark, indicating that there is no ``best'' model, while most of the time, the top-ranked model is simply taken as the absolute winner.}}
    \label{fig:glue_heatmap}
\end{figure}

Figure~\ref{fig:glue_heatmap} shows the performance of different models on different combinations of tasks in terms of their rank in the list of all models. The very top row in the heatmap is the ranking on the SuperGLUE (considering all tasks) and models on the x-axis are sorted based on their rank on ``All'' tasks. There is no strong pattern observable in the plot. For instance, the top ranked model on ``All'' does not newsreel perform best on other combinations of tasks. 

\newpage
\section{SuperGLUE: Rank correlation between SuperGLUE score and individual tasks}
\label{app:superglue_rank_corr}
\begin{wrapfigure}{r}{0.38\textwidth}
    \vspace{-30pt}
  \begin{center}
    \includegraphics[width=0.27\textwidth]{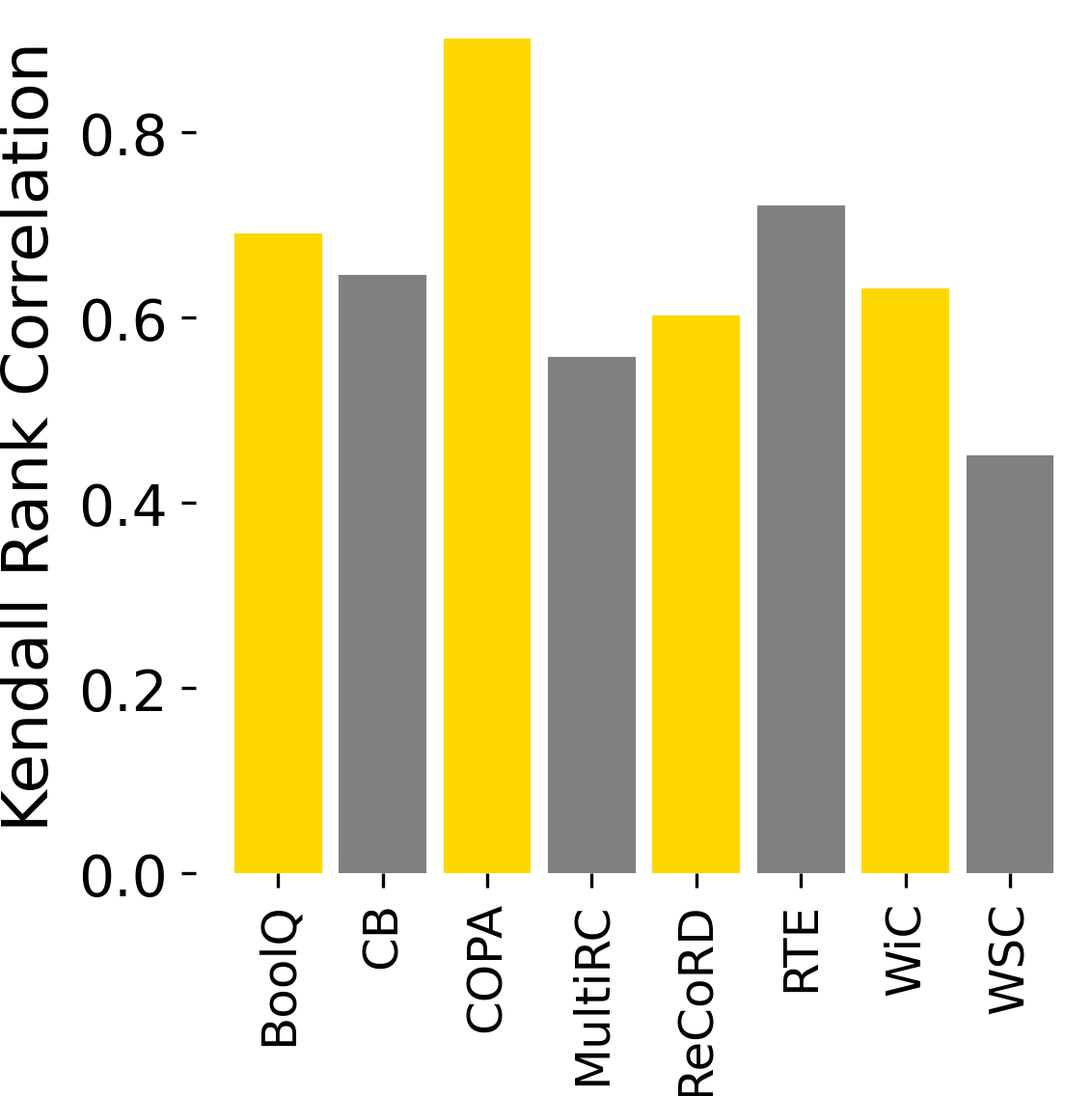}
  \end{center}
  \vspace{-8pt}
  \small
  \caption{ \small{Rank correlation between the SuperGLUE mean score and task's scores.}}
  \label{fig:superglue_rank_corr}
  \vspace{-20pt}
\end{wrapfigure}
Figure~\ref{fig:superglue_rank_corr} shows the rank correlation between the SuperGLUE score and each of the 8 tasks in the benchmark, given 55 different models described in~\citep{narang2021transformer}. The average Kendall rank correlation of tasks with the SuperGLUE score is $0.648$. This correlation is not perfect, but a more important point in the SuperGLUE benchmark is the disagreement of the top-k models across all tasks.  This point is highlighted in Figure~\ref{fig:superglue_unique_rankings}, where for instance, in 6 out of 8 individual tasks, we have different models as the winner. Thus using the mean score for a practitioner to choose a model to adapt it for their own application can be sub-optimal based on the context.

\section{SuperGLUE: Effect on relative ranking of models}
\label{app:superglue_rel_rank}
A significant amount of work in machine learning modeling is to determine to the relative performance of a set of different inductive biases or model architectures. We show that the ranking of models can be drastically altered based on the choice of the subset of the benchmark considered. In other words, the relative performance of models can be easily manipulated by task selection. In order to show this phenomenon, we select ten models, namely vanilla Transformers, Weighted Transformers, Funnel Transformers, Switch Transformers, Lightweight Convolutions, Dynamic Convolutions, Universal Transformers and Adaptive Softmax. The results are similarly obtained from~\citep{narang2021transformer}. 
Table \ref{tab:relative_order} reports a sample of different selection of tasks. We show that for a different selection of tasks, the relative order of model performance is \emph{very} different. Notably, models such as Universal, MoE, Adaptive Softmax, Switch all take turns to become the best performing model on multiple task configurations. Consequently, it is easy to see that with some manipulation of the benchmark configuration, it is easy to endorse and favor the performance of one model over another. 

\begin{table}[h]
    \centering
    \small
    \caption{Relative order of different models when selecting different subsets of SuperGLUE. Selecting different subsets of tasks can produce \emph{very different} outcomes for relative ranking of model architectures. Models that did not appear in Top-5 at all are Lightweight Conv, Dynamic Conv and Transparent Attention. For tasks, A=BoolQ, B=CB, C=CoPA, D=MultiRC, E=ReCoRD, F=RTE, G=WiC, H=WSC.}
    \begin{tabular}{c|l}
    \hline 
      Tasks  & Top-5 Performing Models (In Order) \\
      \hline
      H & Universal, Switch, Adaptive Softmax, Weighted, Vanilla \\
      G & MoE, Switch, Vanilla, Funnel, Universal \\
         A, B& Adaptive Softmax, Vanilla, MoE, Switch, Weighted \\
         A, C & MoE, Switch, Adaptive Softmax, Vanilla, Universal \\
         D, H & Switch, Universal, Adaptive Softmax, MoE, Weighted \\
         B, E, H & Adaptive Softmax, Switch, MoE, Vanilla, Weighted \\
         F, G, H & Switch, MoE, Adaptive Softmax, Universal, Vanilla \\
         A, F, G & MoE, Switch, Vanilla, Adaptive Softmax, Vanilla \\ 
         C, F, G, H & Switch, MoE, Adaptive Softmax, Vanilla, Universal \\ 
         A, C, D, G & MoE, Switch, Adaptive Softmax, Vanilla, Universal \\ 
         \hline 
         All & Switch, MoE, Adaptive Softmax, Vanilla, Universal\\
         \hline
    \end{tabular}
    \label{tab:relative_order}
\end{table}

Optimistically, we also note that even under the notion of a \emph{lottery}, not all models have equal odds. Models that perform poorly across all tasks generally tend to not have a chance to qualify for the Top-5 of any of the above benchmark configurations. We note that Lightweight convolutions, Dynamic Convolutions, and Transparent Attention never made it to any of the Top-5 rankings. Models such as Funnel and Weighted also make very limited appearances. In short, we show empirically that benchmark suites can do pretty well in filtering model architectures that do poorly on most tasks.


\newpage
\section{VTAB: Agreement on top-ranked models across sub-categories and tasks}
\label{app:vtab_agreement}
\begin{figure}[h]
     \centering
     \begin{subfigure}[b]{0.51\textwidth}
         \centering
         \includegraphics[width=\textwidth]{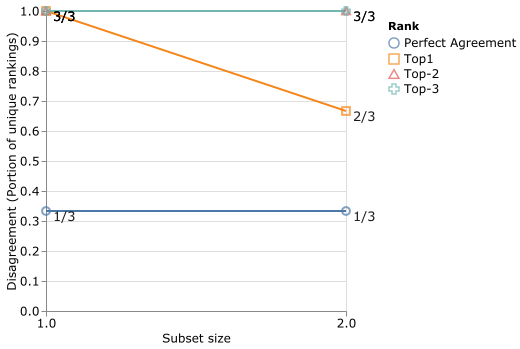}
         \caption{All different subsets of categories.}
         \label{fig:vtab_unique_rankings_subcats}
     \end{subfigure}
     \begin{subfigure}[b]{0.40\textwidth}
         \centering
         \includegraphics[width=\textwidth]{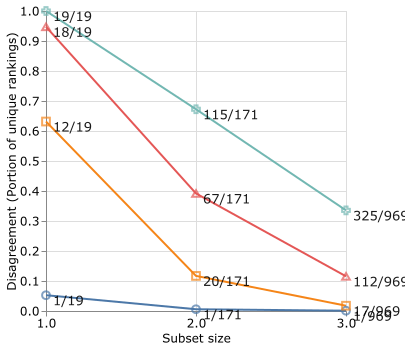}
         \caption{Subsets (with size 1, 2, and 3) of tasks.}
         \label{fig:vtab_unique_rankings_tasks}
     \end{subfigure}
    \caption{Disagreement of model rankings on the VTAB benchmark as a function of the number of selected benchmark sub-categories (3 sub-categories: Natural, Specialized, Structured) or tasks (19 different tasks).}
    \label{fig:vtab_unique_rankings}
\end{figure}

Similar to Figure~\ref{fig:superglue_unique_rankings}, we looked into the disagreement of Top-1, 2, and 3 models based on different combinations of three VTAB sub-categories as well as different combinations with sizes 1, 2, and 3 of VTAB tasks. The $x$-axis represents the number of sub-categories/tasks in each sub-selection and each line corresponds to a different value of $k$ for the Top-$k$ in the rankings. Points are labels as $A / B$, where $A$ is the number of unique model rankings and $B$ is the total number of possible sub-category combinations for this subset size. 

In Figure~\ref{fig:vtab_unique_rankings_subcats}, we can see that all categories disagree on the wining model (top-1) and there is no full agreement on the set of top-2 or top-3 models. We can see a similar disagreement between subsets of tasks in Figure~\ref{fig:vtab_unique_rankings_tasks}. For instance, out of 19 individual tasks (the subset of size 1), there are 12 different winners (top-1) model or looking at the subsets of size 2 tasks, there are 20 different winners. Note that although the disagreement portion is 20/171 is rather small, at the end of the day, we have 20 different models performing best in different situations, and taking a single model based on the VTAB score as the best one can be easily a sub-optimal choice for many scenarios. 

\newpage
\section{LRA: Comparing the top-scoring models on all possible task subsets}
\label{app:lra_all}
Table~\ref{tab:lra-full} presents the Top-1, 2, and 3 ranked results based on all possible combinations of tasks in Long Range Arena benchmark. This in fact is the complete version of Table~\ref{tab:lra} in Section~\ref{sec:lra}.

\begin{table}[h]
    \centering
    \small
    \caption{Top 3 performing models on LRA depending on which subset of tasks we select.}
    \label{tab:lra-full}
    \begin{tabular}{l|ccc}
    \hline
    Task & Best Model & Rank-2 & Rank-3 \\
    \hline
 $t_1$(Text only)       & Linear Transformers & Performer & Transformer \\ 
 $t_2$(Retrieval only) & Sparse Transformers & BigBird & Longformer \\
 $t_3$(ListOps only) & Reformer &  Synthesizer & Transformer \\
 $t_4$ (Image only) &  Sparse Transformer & Performer & Transformer \\
 $t_5$(Path only) & Performer & Linformer & Linear Transformers \\ 
 \hline
$t_1$+ $t_2$& BigBird & Sparse Transformer & Transformer \\
$t_1$+ $t_3$& Transformer & BigBird & Synthesizer\\
$t_1$+ $t_4$ & Linear Transformer & Performer & Transformer \\
$t_1$+ $t_5$& Performer & BigBird & Transformer \\
$t_2$+ $t_3$& BigBird & Transformer & Longformer\\ 
$t_2$+ $t_4$ & Sparse Transformer & BigBird & Transformer\\
$t_2$+ $t_5$& BigBird & Sparse Transformer &  Performer \\
$t_3$+ $t_5$& Linformer & BigBird & Transformer\\
$t_3$+ $t_4$ & Transformer & Synthesizer & Longformer\\ 
$t_4$ + $t_5$& Performer & Linear Transformer & Sparse Transformer\\
\hline
$t_1$+ $t_2$+ $t_3$& BigBird & Transformer & Synthesizer\\ 
$t_1$+ $t_2$+ $t_4$ & Sparse Transformer & Transformer & BigBird \\
$t_1$+ $t_2$+ $t_5$& Performer & Linear Transformer & Transformer \\ 
$t_2$+ $t_3$+ $t_4$ & Transformer & Longformer & Synthesizer\\
$t_2$+ $t_3$+ $t_5$& BigBird & Transformer & Longformer\\
$t_3$+ $t_4$ + $t_5$& BigBird & Transformer & Longformer \\
\hline
$t_1$+ $t_2$+ $t_3$+ $t_4$ & Transformer & BigBird & Longformer\\
$t_1$+ $t_3$+ $t_4$ + $t_5$& BigBird & Transformer & Longformer\\
$t_1$+ $t_2$+ $t_4$ + $t_5$& Sparse Transformer & Performer & BigBird\\ 
$t_2$+ $t_3$+ $t_4$ + $t_5$& BigBird & Transformer & Longformer\\
\hline
$t_1$+ $t_2$+ $t_3$+ $t_4$ + $t_5$ (LRA Score) & BigBird & Transformer & Longformer\\
 \hline
    \end{tabular}
\end{table}

\clearpage
\newpage
\section{Popular public benchmarks for evaluating recommend systems}
\label{app:rec_sys}
Table~\ref{tab:recsys} presets the list of publicly available datasets for recommender systems used by the community for evaluation.  
\begin{savenotes}
\begin{table}[!htp]
    \centering
        \small
        \caption{List of popular offline datasets used for evaulating recommender systems.}
    \label{tab:recsys}
    \begin{tabular}{c|cccc}
    \hline
         Dataset & Number of examples & Users & Items & Sparsity \\
        \hline
      MovieLens 1M \footnote{\url{https://grouplens.org/datasets/movielens/1m/}} & 1,000,209 & 3706 & 6040 & 95.53\%\\
      Movielens 20M \footnote{\url{https://grouplens.org/datasets/movielens/20m/}} & 13,501,622 & 138,159 & 16,954 & 99.42\%\\
      Amazon Product Review (Movies \& TV) \footnote{\url{https://jmcauley.ucsd.edu/data/amazon/}} & 505K & 22,147 & 178,086 & 99.98\%\\
      Amazon Product Review (Video Games) & 46K & 2,670 & 47,063 & 99.96\%\\
      Yahoo Movies \footnote{\url{https://webscope.sandbox.yahoo.com/catalog.php?datatype=r}} & 221,367 & 7,642 & 11,915 & 99.76\%\\
      Pinterest \footnote{\url{https://paperswithcode.com/dataset/pinterest}} & 1.5M & 9916 & 55187 & 99.73\%\\
      Xing \footnote{\url{http://www.recsyschallenge.com/2017/}} & 1,450,300 & 65,347 & 20,778 & 99.89\%\\
      Taobao \footnote{\url{https://tianchi.aliyun.com/dataset/dataDetail?dataId=649}} & 100M & 968K & 4M & 99.98\%\\
      Last.FM \footnote{\url{http://ocelma.net/MusicRecommendationDataset/}} & 42,346 & 1,872 & 3,846 & 99.41\%\\
      Book-Crossing \footnote{\url{http://www2.informatik.uni-freiburg.de/~cziegler/BX/}} & 172,576 & 19,676 & 20,003 & 99.96\%\\
          \hline
    \end{tabular}
\end{table}
\end{savenotes}

\clearpage
\newpage
\section{Benchmarking checklist for the review process}
\label{app:checklist}
This section presents a proposal for a checklist that can be used in the review process with the hope of reducing the benchmark lottery effect. 
Note that as we discussed in the paper, the benchmark lottery effect can be rooted in various aspects. As an example, \citet{gebru2018datasheets} provided a list of questions for the benchmark creation process that covers motivation of the benchmark, composition, collection process, and recommended uses. Such data can be also framed as checklists for benchmark creation.

\begin{tcolorbox}[colback=white!5,colframe=white!40!black,title=Benchmarking checklist for reviewers and area chairs]
\begin{itemize}
\renewcommand\labelitemi{$\square$}
    
    \item If there is written dissatisfaction about the author's choice of baselines, tasks, or benchmarks in the reviews, are there rationals beyond the fact that these requested datasets are ``must-have'' benchmarks?
        
    \item Are the reviews considering potential benefits like efficiency, fairness, and simplicity of the proposed model outside the commonly evaluated performance metrics (e.g., accuracy)?
    
    \item Are there any negative points in the reviews due to the paper proposing a method that deviates from the current trend/hype. If so, are there rational justifications for this?
    
    \item If the reviews penalizing the paper due to the proposed method not performing well only on a subset of tasks, is there enough logical elaboration on such criticism in the reviews? 
    
    \item Are the reviews assessing the evaluation strategy in terms of studying the effect of different sources of variance (e.g., multiple splits, multiple random seeds, etc.)?
    
    \item If there are analyses on statistical significance testing, are they appreciated in the reviews? If there is no such analysis, are there recommendations on this provided in the reviews? 

    \item If the paper is claiming SOTA or improvements over baselines on a benchmark, are there ablations on how much such improvement is secured by the tricks that are not tied to the main contributions? 
    
    \item If the reviews are asking for more experiments, analysis, or evaluation on more benchmarks, are the potential blockers are considered for such requests? E.g. those experiments being out of reach in terms of computing budget (pre-training or extremely large datasets).
    
    \item If the paper is proposing a new idea while deviating from the common paradigms, is the ``out of the hype'' thinking valued in the reviews as opposed to solely recognizing SOTA performance?
    
\end{itemize}
\end{tcolorbox}

%% file: main.bbl
\begin{thebibliography}{148}
\providecommand{\natexlab}[1]{#1}
\providecommand{\url}[1]{\texttt{#1}}
\expandafter\ifx\csname urlstyle\endcsname\relax
  \providecommand{\doi}[1]{doi: #1}\else
  \providecommand{\doi}{doi: \begingroup \urlstyle{rm}\Url}\fi

\bibitem[Abnar et~al.(2020)Abnar, Dehghani, and Zuidema]{abnar2020transferring}
Samira Abnar, Mostafa Dehghani, and Willem Zuidema.
\newblock Transferring inductive biases through knowledge distillation.
\newblock \emph{arXiv preprint arXiv:2006.00555}, 2020.

\bibitem[Albrecht et~al.(2015)Albrecht, Beck, Buckeridge, Botea, Caragea, Chi,
  Damoulas, Dilkina, Eaton, Fazli, et~al.]{albrecht2015reports}
Stefano~V Albrecht, J~Christopher Beck, David~L Buckeridge, Adi Botea, Cornelia
  Caragea, Chi-hung Chi, Theodoros Damoulas, Bistra Dilkina, Eric Eaton, Pooyan
  Fazli, et~al.
\newblock Reports on the 2015 aaai workshop program.
\newblock \emph{Ai Magazine}, 36\penalty0 (2):\penalty0 90--101, 2015.

\bibitem[Amidei et~al.(2018)Amidei, Piwek, and Willis]{amidei2018evaluation}
Jacopo Amidei, Paul Piwek, and Alistair Willis.
\newblock Evaluation methodologies in automatic question generation 2013-2018.
\newblock 2018.

\bibitem[Amidei et~al.(2020)Amidei, Piwek, and Willis]{amidei2020identifying}
Jacopo Amidei, Paul Piwek, and Alistair Willis.
\newblock Identifying annotator bias: A new irt-based method for bias
  identification.
\newblock 2020.

\bibitem[Armstrong et~al.(2009)Armstrong, Moffat, Webber, and
  Zobel]{armstrong2009improvements}
Timothy~G Armstrong, Alistair Moffat, William Webber, and Justin Zobel.
\newblock Improvements that don't add up: ad-hoc retrieval results since 1998.
\newblock In \emph{Proceedings of the 18th ACM conference on Information and
  knowledge management}, pages 601--610, 2009.

\bibitem[Arora and Zhang(2021)]{arora2021rip}
Sanjeev Arora and Yi~Zhang.
\newblock Rip van winkle's razor: A simple estimate of overfit to test data.
\newblock \emph{arXiv preprint arXiv:2102.13189}, 2021.

\bibitem[Bahdanau et~al.(2014)Bahdanau, Cho, and Bengio]{bahdanau2014neural}
Dzmitry Bahdanau, Kyunghyun Cho, and Yoshua Bengio.
\newblock Neural machine translation by jointly learning to align and
  translate.
\newblock \emph{arXiv preprint arXiv:1409.0473}, 2014.

\bibitem[Balduzzi et~al.(2018)Balduzzi, Tuyls, Perolat, and
  Graepel]{balduzzi2018re}
David Balduzzi, Karl Tuyls, Julien Perolat, and Thore Graepel.
\newblock Re-evaluating evaluation.
\newblock \emph{arXiv preprint arXiv:1806.02643}, 2018.

\bibitem[Beattie et~al.(2016)Beattie, Leibo, Teplyashin, Ward, Wainwright,
  K{\"u}ttler, Lefrancq, Green, Vald{\'e}s, Sadik, et~al.]{beattie2016deepmind}
Charles Beattie, Joel~Z Leibo, Denis Teplyashin, Tom Ward, Marcus Wainwright,
  Heinrich K{\"u}ttler, Andrew Lefrancq, Simon Green, V{\'\i}ctor Vald{\'e}s,
  Amir Sadik, et~al.
\newblock Deepmind lab.
\newblock \emph{arXiv preprint arXiv:1612.03801}, 2016.

\bibitem[Bellemare et~al.(2013)Bellemare, Naddaf, Veness, and
  Bowling]{bellemare2013arcade}
Marc~G Bellemare, Yavar Naddaf, Joel Veness, and Michael Bowling.
\newblock The arcade learning environment: An evaluation platform for general
  agents.
\newblock \emph{Journal of Artificial Intelligence Research}, 47:\penalty0
  253--279, 2013.

\bibitem[Bello et~al.(2021)Bello, Fedus, Du, Cubuk, Srinivas, Lin, Shlens, and
  Zoph]{bello2021revisiting}
Irwan Bello, William Fedus, Xianzhi Du, Ekin~D Cubuk, Aravind Srinivas,
  Tsung-Yi Lin, Jonathon Shlens, and Barret Zoph.
\newblock Revisiting resnets: Improved training and scaling strategies.
\newblock \emph{arXiv preprint arXiv:2103.07579}, 2021.

\bibitem[Bender and Friedman(2018)]{bender2018data}
Emily~M Bender and Batya Friedman.
\newblock Data statements for natural language processing: Toward mitigating
  system bias and enabling better science.
\newblock \emph{Transactions of the Association for Computational Linguistics},
  6:\penalty0 587--604, 2018.

\bibitem[Beutel et~al.(2017)Beutel, Chi, Cheng, Pham, and
  Anderson]{beutel2017beyond}
Alex Beutel, Ed~H Chi, Zhiyuan Cheng, Hubert Pham, and John Anderson.
\newblock Beyond globally optimal: Focused learning for improved
  recommendations.
\newblock In \emph{Proceedings of the 26th International Conference on World
  Wide Web}, pages 203--212, 2017.

\bibitem[Beyer et~al.(2020)Beyer, H{\'e}naff, Kolesnikov, Zhai, and
  Oord]{beyer2020we}
Lucas Beyer, Olivier~J H{\'e}naff, Alexander Kolesnikov, Xiaohua Zhai, and
  A{\"a}ron van~den Oord.
\newblock Are we done with {ImageNet}?
\newblock \emph{arXiv preprint arXiv:2006.07159}, 2020.

\bibitem[Blum and Hardt(2015)]{blum2015ladder}
Avrim Blum and Moritz Hardt.
\newblock The ladder: A reliable leaderboard for machine learning competitions.
\newblock In \emph{International Conference on Machine Learning}, pages
  1006--1014. PMLR, 2015.

\bibitem[Bouthillier et~al.(2019)Bouthillier, Laurent, and
  Vincent]{bouthillier2019unreproducible}
Xavier Bouthillier, C{\'e}sar Laurent, and Pascal Vincent.
\newblock Unreproducible research is reproducible.
\newblock In \emph{International Conference on Machine Learning}, pages
  725--734. PMLR, 2019.

\bibitem[Bouthillier et~al.(2021)Bouthillier, Delaunay, Bronzi, Trofimov,
  Nichyporuk, Szeto, Mohammadi~Sepahvand, Raff, Madan, Voleti,
  et~al.]{Bouthillier2021}
Xavier Bouthillier, Pierre Delaunay, Mirko Bronzi, Assya Trofimov, Brennan
  Nichyporuk, Justin Szeto, Nazanin Mohammadi~Sepahvand, Edward Raff, Kanika
  Madan, Vikram Voleti, et~al.
\newblock Accounting for variance in machine learning benchmarks.
\newblock \emph{Proceedings of Machine Learning and Systems}, 3, 2021.

\bibitem[Bowman and Dahl(2021)]{bowman2021will}
Samuel~R Bowman and George~E Dahl.
\newblock What will it take to fix benchmarking in natural language
  understanding?
\newblock \emph{arXiv preprint arXiv:2104.02145}, 2021.

\bibitem[Bowman et~al.(2015)Bowman, Angeli, Potts, and
  Manning]{bowman2015large}
Samuel~R Bowman, Gabor Angeli, Christopher Potts, and Christopher~D Manning.
\newblock A large annotated corpus for learning natural language inference.
\newblock \emph{arXiv preprint arXiv:1508.05326}, 2015.

\bibitem[Brown et~al.(2020)Brown, Mann, Ryder, Subbiah, Kaplan, Dhariwal,
  Neelakantan, Shyam, Sastry, Askell, et~al.]{brown2020language}
Tom~B Brown, Benjamin Mann, Nick Ryder, Melanie Subbiah, Jared Kaplan, Prafulla
  Dhariwal, Arvind Neelakantan, Pranav Shyam, Girish Sastry, Amanda Askell,
  et~al.
\newblock Language models are few-shot learners.
\newblock \emph{arXiv preprint arXiv:2005.14165}, 2020.

\bibitem[Cheng et~al.(2017)Cheng, Han, and Lu]{cheng2017remote}
Gong Cheng, Junwei Han, and Xiaoqiang Lu.
\newblock Remote sensing image scene classification: Benchmark and state of the
  art.
\newblock \emph{Proceedings of the IEEE}, 105\penalty0 (10):\penalty0
  1865--1883, 2017.

\bibitem[Cimpoi et~al.(2014)Cimpoi, Maji, Kokkinos, Mohamed, and
  Vedaldi]{cimpoi2014describing}
Mircea Cimpoi, Subhransu Maji, Iasonas Kokkinos, Sammy Mohamed, and Andrea
  Vedaldi.
\newblock Describing textures in the wild.
\newblock In \emph{Proceedings of the IEEE Conference on Computer Vision and
  Pattern Recognition}, pages 3606--3613, 2014.

\bibitem[Clark et~al.(2020)Clark, Luong, Le, and Manning]{clark2020electra}
Kevin Clark, Minh-Thang Luong, Quoc~V Le, and Christopher~D Manning.
\newblock Electra: Pre-training text encoders as discriminators rather than
  generators.
\newblock \emph{arXiv preprint arXiv:2003.10555}, 2020.

\bibitem[Dacrema et~al.(2019)Dacrema, Cremonesi, and Jannach]{dacrema2019we}
Maurizio~Ferrari Dacrema, Paolo Cremonesi, and Dietmar Jannach.
\newblock Are we really making much progress? a worrying analysis of recent
  neural recommendation approaches.
\newblock In \emph{Proceedings of the 13th ACM Conference on Recommender
  Systems}, pages 101--109, 2019.

\bibitem[Dehghani et~al.(2019)Dehghani, Gouws, Vinyals, Uszkoreit, and
  Kaiser]{universaltrans}
Mostafa Dehghani, Stephan Gouws, Oriol Vinyals, Jakob Uszkoreit, and Lukasz
  Kaiser.
\newblock Universal transformers.
\newblock In \emph{Proceedings of the 7th International Conference on Learning
  Representations}, ICLR'19, 2019.
\newblock URL \url{https://arxiv.org/abs/1807.03819}.

\bibitem[Dem{\v{s}}ar(2006)]{Demvsar2006}
Janez Dem{\v{s}}ar.
\newblock Statistical comparisons of classifiers over multiple data sets.
\newblock \emph{The Journal of Machine Learning Research}, 7:\penalty0 1--30,
  2006.

\bibitem[Deng et~al.(2009)Deng, Dong, Socher, Li, Li, and
  Fei-Fei]{deng2009imagenet}
Jia Deng, Wei Dong, Richard Socher, Li-Jia Li, Kai Li, and Li~Fei-Fei.
\newblock {ImageNet}: {A} large-scale hierarchical image database.
\newblock In \emph{2009 IEEE conference on computer vision and pattern
  recognition}, pages 248--255. Ieee, 2009.

\bibitem[Denton et~al.(2020)Denton, Hanna, Amironesei, Smart, Nicole, and
  Scheuerman]{denton2020bringing}
Emily Denton, Alex Hanna, Razvan Amironesei, Andrew Smart, Hilary Nicole, and
  Morgan~Klaus Scheuerman.
\newblock Bringing the people back in: Contesting benchmark machine learning
  datasets.
\newblock \emph{arXiv preprint arXiv:2007.07399}, 2020.

\bibitem[Devlin et~al.(2018)Devlin, Chang, Lee, and Toutanova]{devlin2018bert}
Jacob Devlin, Ming-Wei Chang, Kenton Lee, and Kristina Toutanova.
\newblock Bert: Pre-training of deep bidirectional transformers for language
  understanding.
\newblock \emph{arXiv preprint arXiv:1810.04805}, 2018.

\bibitem[Dodge et~al.(2019)Dodge, Gururangan, Card, Schwartz, and
  Smith]{dodge2019show}
Jesse Dodge, Suchin Gururangan, Dallas Card, Roy Schwartz, and Noah~A Smith.
\newblock Show your work: Improved reporting of experimental results.
\newblock \emph{arXiv preprint arXiv:1909.03004}, 2019.

\bibitem[Dosovitskiy et~al.(2020)Dosovitskiy, Beyer, Kolesnikov, Weissenborn,
  Zhai, Unterthiner, Dehghani, Minderer, Heigold, Gelly,
  et~al.]{dosovitskiy2020image}
Alexey Dosovitskiy, Lucas Beyer, Alexander Kolesnikov, Dirk Weissenborn,
  Xiaohua Zhai, Thomas Unterthiner, Mostafa Dehghani, Matthias Minderer, Georg
  Heigold, Sylvain Gelly, et~al.
\newblock An image is worth 16x16 words: Transformers for image recognition at
  scale.
\newblock \emph{arXiv preprint arXiv:2010.11929}, 2020.

\bibitem[Dror et~al.(2017)Dror, Baumer, Bogomolov, and Reichart]{Dror2017}
Rotem Dror, Gili Baumer, Marina Bogomolov, and Roi Reichart.
\newblock Replicability analysis for natural language processing: Testing
  significance with multiple datasets.
\newblock \emph{Transactions of the Association for Computational Linguistics},
  5:\penalty0 471--486, 2017.

\bibitem[Dulac-Arnold et~al.(2019)Dulac-Arnold, Mankowitz, and
  Hester]{dulac2019challenges}
Gabriel Dulac-Arnold, Daniel Mankowitz, and Todd Hester.
\newblock Challenges of real-world reinforcement learning.
\newblock \emph{arXiv preprint arXiv:1904.12901}, 2019.

\bibitem[Dwork et~al.(2001)Dwork, Kumar, Naor, and Sivakumar]{dwork2001rank}
Cynthia Dwork, Ravi Kumar, Moni Naor, and Dandapani Sivakumar.
\newblock Rank aggregation methods for the web.
\newblock In \emph{Proceedings of the 10th international conference on World
  Wide Web}, pages 613--622, 2001.

\bibitem[Dwork et~al.(2015)Dwork, Feldman, Hardt, Pitassi, Reingold, and
  Roth]{dwork2015reusable}
Cynthia Dwork, Vitaly Feldman, Moritz Hardt, Toniann Pitassi, Omer Reingold,
  and Aaron Roth.
\newblock The reusable holdout: Preserving validity in adaptive data analysis.
\newblock \emph{Science}, 349\penalty0 (6248):\penalty0 636--638, 2015.

\bibitem[Ethayarajh and Jurafsky(2020)]{ethayarajh2020utility}
Kawin Ethayarajh and Dan Jurafsky.
\newblock Utility is in the eye of the user: A critique of nlp leaderboards.
\newblock \emph{arXiv preprint arXiv:2009.13888}, 2020.

\bibitem[Fedus et~al.(2021)Fedus, Zoph, and Shazeer]{fedus2021switch}
William Fedus, Barret Zoph, and Noam Shazeer.
\newblock Switch transformers: Scaling to trillion parameter models with simple
  and efficient sparsity.
\newblock \emph{arXiv preprint arXiv:2101.03961}, 2021.

\bibitem[Fei-Fei et~al.(2006)Fei-Fei, Fergus, and Perona]{fei2006one}
Li~Fei-Fei, Rob Fergus, and Pietro Perona.
\newblock One-shot learning of object categories.
\newblock \emph{IEEE transactions on pattern analysis and machine
  intelligence}, 28\penalty0 (4):\penalty0 594--611, 2006.

\bibitem[Feldman et~al.(2019{\natexlab{a}})Feldman, Frostig, and
  Hardt]{feldman2019advantages}
Vitaly Feldman, Roy Frostig, and Moritz Hardt.
\newblock The advantages of multiple classes for reducing overfitting from test
  set reuse.
\newblock In \emph{International Conference on Machine Learning}, pages
  1892--1900. PMLR, 2019{\natexlab{a}}.

\bibitem[Feldman et~al.(2019{\natexlab{b}})Feldman, Frostig, and
  Hardt]{feldman2019open}
Vitaly Feldman, Roy Frostig, and Moritz Hardt.
\newblock Open problem: How fast can a multiclass test set be overfit?
\newblock In \emph{Conference on Learning Theory}, pages 3185--3189. PMLR,
  2019{\natexlab{b}}.

\bibitem[Gebru et~al.(2018)Gebru, Morgenstern, Vecchione, Vaughan, Wallach,
  Daum{\'e}~III, and Crawford]{gebru2018datasheets}
Timnit Gebru, Jamie Morgenstern, Briana Vecchione, Jennifer~Wortman Vaughan,
  Hanna Wallach, Hal Daum{\'e}~III, and Kate Crawford.
\newblock Datasheets for datasets.
\newblock \emph{arXiv preprint arXiv:1803.09010}, 2018.

\bibitem[Gehrmann et~al.(2021)Gehrmann, Adewumi, Aggarwal, Ammanamanchi,
  Anuoluwapo, Bosselut, Chandu, Clinciu, Das, Dhole, et~al.]{gehrmann2021gem}
Sebastian Gehrmann, Tosin Adewumi, Karmanya Aggarwal, Pawan~Sasanka
  Ammanamanchi, Aremu Anuoluwapo, Antoine Bosselut, Khyathi~Raghavi Chandu,
  Miruna Clinciu, Dipanjan Das, Kaustubh~D Dhole, et~al.
\newblock The {GEM} benchmark: Natural language generation, its evaluation and
  metrics.
\newblock \emph{arXiv preprint arXiv:2102.01672}, 2021.

\bibitem[Geiger et~al.(2013)Geiger, Lenz, Stiller, and
  Urtasun]{geiger2013vision}
Andreas Geiger, Philip Lenz, Christoph Stiller, and Raquel Urtasun.
\newblock Vision meets robotics: The kitti dataset.
\newblock \emph{The International Journal of Robotics Research}, 32\penalty0
  (11):\penalty0 1231--1237, 2013.

\bibitem[Goodfellow et~al.(2014)Goodfellow, Pouget-Abadie, Mirza, Xu,
  Warde-Farley, Ozair, Courville, and Bengio]{goodfellow2014generative}
Ian Goodfellow, Jean Pouget-Abadie, Mehdi Mirza, Bing Xu, David Warde-Farley,
  Sherjil Ozair, Aaron Courville, and Yoshua Bengio.
\newblock Generative adversarial nets.
\newblock \emph{Advances in neural information processing systems}, 27, 2014.

\bibitem[Gulcehre et~al.(2020)Gulcehre, Wang, Novikov, Paine, G{\'o}mez, Zolna,
  Agarwal, Merel, Mankowitz, Paduraru, et~al.]{gulcehre2020rl}
Caglar Gulcehre, Ziyu Wang, Alexander Novikov, Thomas Paine, Sergio G{\'o}mez,
  Konrad Zolna, Rishabh Agarwal, Josh~S Merel, Daniel~J Mankowitz, Cosmin
  Paduraru, et~al.
\newblock Rl unplugged: A collection of benchmarks for offline reinforcement
  learning.
\newblock \emph{Advances in Neural Information Processing Systems}, 33, 2020.

\bibitem[Guo et~al.(2019)Guo, Geng, Simcha, Chern, Kumar, and
  Wu]{DBLP:journals/corr/abs-1908-10396}
Ruiqi Guo, Quan Geng, David Simcha, Felix Chern, Sanjiv Kumar, and Xiang Wu.
\newblock New loss functions for fast maximum inner product search.
\newblock \emph{CoRR}, abs/1908.10396, 2019.
\newblock URL \url{http://arxiv.org/abs/1908.10396}.

\bibitem[Gururangan et~al.(2018)Gururangan, Swayamdipta, Levy, Schwartz,
  Bowman, and Smith]{gururangan-etal-2018-annotation}
Suchin Gururangan, Swabha Swayamdipta, Omer Levy, Roy Schwartz, Samuel Bowman,
  and Noah~A. Smith.
\newblock Annotation artifacts in natural language inference data.
\newblock In \emph{Proceedings of the 2018 Conference of the North {A}merican
  Chapter of the Association for Computational Linguistics: Human Language
  Technologies, Volume 2 (Short Papers)}, pages 107--112, New Orleans,
  Louisiana, June 2018. Association for Computational Linguistics.
\newblock \doi{10.18653/v1/N18-2017}.
\newblock URL \url{https://www.aclweb.org/anthology/N18-2017}.

\bibitem[Hao and Ryan(2016)]{hao2016real}
Feng Hao and Peter~YA Ryan.
\newblock \emph{Real-world electronic voting: Design, analysis and deployment}.
\newblock CRC Press, 2016.

\bibitem[Harper and Konstan(2015)]{10.1145/2827872}
F.~Maxwell Harper and Joseph~A. Konstan.
\newblock The movielens datasets: History and context.
\newblock \emph{ACM Trans. Interact. Intell. Syst.}, 5\penalty0 (4), December
  2015.
\newblock ISSN 2160-6455.
\newblock \doi{10.1145/2827872}.
\newblock URL \url{https://doi.org/10.1145/2827872}.

\bibitem[Hausknecht et~al.(2014)Hausknecht, Lehman, Miikkulainen, and
  Stone]{hausknecht2014neuroevolution}
Matthew Hausknecht, Joel Lehman, Risto Miikkulainen, and Peter Stone.
\newblock A neuroevolution approach to general atari game playing.
\newblock \emph{IEEE Transactions on Computational Intelligence and AI in
  Games}, 6\penalty0 (4):\penalty0 355--366, 2014.

\bibitem[He et~al.(2016{\natexlab{a}})He, Zhang, Ren, and Sun]{he2016deep}
Kaiming He, Xiangyu Zhang, Shaoqing Ren, and Jian Sun.
\newblock Deep residual learning for image recognition.
\newblock In \emph{Proceedings of the IEEE conference on computer vision and
  pattern recognition}, pages 770--778, 2016{\natexlab{a}}.

\bibitem[He and McAuley(2016)]{he2016ups}
Ruining He and Julian McAuley.
\newblock Ups and downs: Modeling the visual evolution of fashion trends with
  one-class collaborative filtering.
\newblock In \emph{proceedings of the 25th international conference on world
  wide web}, pages 507--517, 2016.

\bibitem[He et~al.(2016{\natexlab{b}})He, Zhang, Kan, and Chua]{he2016fast}
Xiangnan He, Hanwang Zhang, Min-Yen Kan, and Tat-Seng Chua.
\newblock Fast matrix factorization for online recommendation with implicit
  feedback.
\newblock In \emph{Proceedings of the 39th International ACM SIGIR conference
  on Research and Development in Information Retrieval}, pages 549--558,
  2016{\natexlab{b}}.

\bibitem[He et~al.(2017)He, Liao, Zhang, Nie, Hu, and Chua]{he2017neural}
Xiangnan He, Lizi Liao, Hanwang Zhang, Liqiang Nie, Xia Hu, and Tat-Seng Chua.
\newblock Neural collaborative filtering.
\newblock In \emph{Proceedings of the 26th international conference on world
  wide web}, pages 173--182, 2017.

\bibitem[Heess et~al.(2017)Heess, TB, Sriram, Lemmon, Merel, Wayne, Tassa,
  Erez, Wang, Eslami, et~al.]{heess2017emergence}
Nicolas Heess, Dhruva TB, Srinivasan Sriram, Jay Lemmon, Josh Merel, Greg
  Wayne, Yuval Tassa, Tom Erez, Ziyu Wang, SM~Eslami, et~al.
\newblock Emergence of locomotion behaviours in rich environments.
\newblock \emph{arXiv preprint arXiv:1707.02286}, 2017.

\bibitem[Helber et~al.(2019)Helber, Bischke, Dengel, and
  Borth]{helber2019eurosat}
Patrick Helber, Benjamin Bischke, Andreas Dengel, and Damian Borth.
\newblock Eurosat: A novel dataset and deep learning benchmark for land use and
  land cover classification.
\newblock \emph{IEEE Journal of Selected Topics in Applied Earth Observations
  and Remote Sensing}, 12\penalty0 (7):\penalty0 2217--2226, 2019.

\bibitem[Henderson et~al.(2018)Henderson, Islam, Bachman, Pineau, Precup, and
  Meger]{henderson2018deep}
Peter Henderson, Riashat Islam, Philip Bachman, Joelle Pineau, Doina Precup,
  and David Meger.
\newblock Deep reinforcement learning that matters.
\newblock In \emph{Proceedings of the AAAI Conference on Artificial
  Intelligence}, volume~32, 2018.

\bibitem[Hendrycks and Dietterich(2019)]{hendrycks2019robustness}
Dan Hendrycks and Thomas Dietterich.
\newblock Benchmarking neural network robustness to common corruptions and
  perturbations.
\newblock \emph{Proceedings of the International Conference on Learning
  Representations}, 2019.

\bibitem[Hendrycks et~al.(2020)Hendrycks, Basart, Mu, Kadavath, Wang, Dorundo,
  Desai, Zhu, Parajuli, Guo, Song, Steinhardt, and Gilmer]{hendrycks2020many}
Dan Hendrycks, Steven Basart, Norman Mu, Saurav Kadavath, Frank Wang, Evan
  Dorundo, Rahul Desai, Tyler Zhu, Samyak Parajuli, Mike Guo, Dawn Song, Jacob
  Steinhardt, and Justin Gilmer.
\newblock The many faces of robustness: A critical analysis of
  out-of-distribution generalization.
\newblock \emph{arXiv preprint arXiv:2006.16241}, 2020.

\bibitem[Hendrycks et~al.(2021)Hendrycks, Zhao, Basart, Steinhardt, and
  Song]{hendrycks2021nae}
Dan Hendrycks, Kevin Zhao, Steven Basart, Jacob Steinhardt, and Dawn Song.
\newblock Natural adversarial examples.
\newblock \emph{CVPR}, 2021.

\bibitem[Higgins et~al.(2016)Higgins, Matthey, Pal, Burgess, Glorot, Botvinick,
  Mohamed, and Lerchner]{higgins2016beta}
Irina Higgins, Loic Matthey, Arka Pal, Christopher Burgess, Xavier Glorot,
  Matthew Botvinick, Shakir Mohamed, and Alexander Lerchner.
\newblock beta-vae: Learning basic visual concepts with a constrained
  variational framework.
\newblock 2016.

\bibitem[Hinton et~al.(2015)Hinton, Vinyals, and Dean]{hinton2015distilling}
Geoffrey Hinton, Oriol Vinyals, and Jeff Dean.
\newblock Distilling the knowledge in a neural network.
\newblock \emph{arXiv preprint arXiv:1503.02531}, 2015.

\bibitem[Hooker(2020)]{hooker2020hardware}
Sara Hooker.
\newblock The hardware lottery.
\newblock \emph{arXiv preprint arXiv:2009.06489}, 2020.

\bibitem[Hu et~al.(2020)Hu, Ruder, Siddhant, Neubig, Firat, and
  Johnson]{hu2020xtreme}
Junjie Hu, Sebastian Ruder, Aditya Siddhant, Graham Neubig, Orhan Firat, and
  Melvin Johnson.
\newblock Xtreme: A massively multilingual multi-task benchmark for evaluating
  cross-lingual generalization.
\newblock \emph{CoRR}, abs/2003.11080, 2020.

\bibitem[Jaderberg et~al.(2016)Jaderberg, Mnih, Czarnecki, Schaul, Leibo,
  Silver, and Kavukcuoglu]{jaderberg2016reinforcement}
Max Jaderberg, Volodymyr Mnih, Wojciech~Marian Czarnecki, Tom Schaul, Joel~Z
  Leibo, David Silver, and Koray Kavukcuoglu.
\newblock Reinforcement learning with unsupervised auxiliary tasks.
\newblock \emph{arXiv preprint arXiv:1611.05397}, 2016.

\bibitem[Johnson et~al.(2017)Johnson, Hariharan, Van Der~Maaten, Fei-Fei,
  Lawrence~Zitnick, and Girshick]{johnson2017clevr}
Justin Johnson, Bharath Hariharan, Laurens Van Der~Maaten, Li~Fei-Fei,
  C~Lawrence~Zitnick, and Ross Girshick.
\newblock Clevr: A diagnostic dataset for compositional language and elementary
  visual reasoning.
\newblock In \emph{Proceedings of the IEEE Conference on Computer Vision and
  Pattern Recognition}, pages 2901--2910, 2017.

\bibitem[Kaggle and EyePacs(2015)]{kaggle_retinopathy}
Kaggle and EyePacs.
\newblock Kaggle diabetic retinopathy detection., 2015.
\newblock URL \url{https://www.kaggle.com/c/
  diabetic-retinopathy-detection/data.}

\bibitem[Kiela et~al.(2021)Kiela, Bartolo, Nie, Kaushik, Geiger, Wu, Vidgen,
  Prasad, Singh, Ringshia, et~al.]{kiela2021dynabench}
Douwe Kiela, Max Bartolo, Yixin Nie, Divyansh Kaushik, Atticus Geiger,
  Zhengxuan Wu, Bertie Vidgen, Grusha Prasad, Amanpreet Singh, Pratik Ringshia,
  et~al.
\newblock Dynabench: Rethinking benchmarking in nlp.
\newblock \emph{arXiv preprint arXiv:2104.14337}, 2021.

\bibitem[Kingma and Welling(2013)]{kingma2013auto}
Diederik~P Kingma and Max Welling.
\newblock Auto-encoding variational bayes.
\newblock \emph{arXiv preprint arXiv:1312.6114}, 2013.

\bibitem[Kolesnikov et~al.(2019)Kolesnikov, Beyer, Zhai, Puigcerver, Yung,
  Gelly, and Houlsby]{kolesnikov2019big}
Alexander Kolesnikov, Lucas Beyer, Xiaohua Zhai, Joan Puigcerver, Jessica Yung,
  Sylvain Gelly, and Neil Houlsby.
\newblock Big transfer (bit): General visual representation learning.
\newblock \emph{arXiv preprint arXiv:1912.11370}, 2019.

\bibitem[Krizhevsky et~al.(2009)Krizhevsky, Hinton,
  et~al.]{krizhevsky2009learning}
Alex Krizhevsky, Geoffrey Hinton, et~al.
\newblock Learning multiple layers of features from tiny images.
\newblock 2009.

\bibitem[Lan et~al.(2019)Lan, Chen, Goodman, Gimpel, Sharma, and
  Soricut]{lan2019albert}
Zhenzhong Lan, Mingda Chen, Sebastian Goodman, Kevin Gimpel, Piyush Sharma, and
  Radu Soricut.
\newblock Albert: A lite bert for self-supervised learning of language
  representations.
\newblock \emph{arXiv preprint arXiv:1909.11942}, 2019.

\bibitem[LeCun et~al.(2004)LeCun, Huang, and Bottou]{lecun2004learning}
Yann LeCun, Fu~Jie Huang, and Leon Bottou.
\newblock Learning methods for generic object recognition with invariance to
  pose and lighting.
\newblock In \emph{Proceedings of the 2004 IEEE Computer Society Conference on
  Computer Vision and Pattern Recognition, 2004. CVPR 2004.}, volume~2, pages
  II--104. IEEE, 2004.

\bibitem[Li et~al.(2018)Li, Jamieson, Rostamizadeh, Gonina, Hardt, Recht, and
  Talwalkar]{li2018system}
Liam Li, Kevin Jamieson, Afshin Rostamizadeh, Ekaterina Gonina, Moritz Hardt,
  Benjamin Recht, and Ameet Talwalkar.
\newblock A system for massively parallel hyperparameter tuning.
\newblock \emph{arXiv preprint arXiv:1810.05934}, 2018.

\bibitem[Liang et~al.(2015)Liang, Machado, Talvitie, and
  Bowling]{liang2015state}
Yitao Liang, Marlos~C Machado, Erik Talvitie, and Michael Bowling.
\newblock State of the art control of atari games using shallow reinforcement
  learning.
\newblock \emph{arXiv preprint arXiv:1512.01563}, 2015.

\bibitem[Lin(2019)]{lin2019neural}
Jimmy Lin.
\newblock The neural hype and comparisons against weak baselines.
\newblock \emph{ACM SIGIR Forum}, 52\penalty0 (2):\penalty0 40--51, 2019.

\bibitem[Lin et~al.(2021)Lin, Campos, Craswell, Mitra, and
  Yilmaz]{lin2021significant}
Jimmy Lin, Daniel Campos, Nick Craswell, Bhaskar Mitra, and Emine Yilmaz.
\newblock Significant improvements over the state of the art? a case study of
  the ms marco document ranking leaderboard.
\newblock \emph{arXiv preprint arXiv:2102.12887}, 2021.

\bibitem[Linzen et~al.(2016)Linzen, Dupoux, and Goldberg]{linzen2016assessing}
Tal Linzen, Emmanuel Dupoux, and Yoav Goldberg.
\newblock Assessing the ability of lstms to learn syntax-sensitive
  dependencies.
\newblock \emph{Transactions of the Association for Computational Linguistics},
  4\penalty0 (0), 2016.

\bibitem[Lipovetzky et~al.(2015)Lipovetzky, Ramirez, and
  Geffner]{lipovetzky2015classical}
Nir Lipovetzky, Miquel Ramirez, and Hector Geffner.
\newblock Classical planning with simulators: Results on the atari video games.
\newblock In \emph{Proc. IJCAI}, 2015.

\bibitem[Lipton and Steinhardt(2018)]{lipton2018troubling}
Zachary~C Lipton and Jacob Steinhardt.
\newblock Troubling trends in machine learning scholarship.
\newblock \emph{arXiv preprint arXiv:1807.03341}, 2018.

\bibitem[Liu et~al.(2019)Liu, Ott, Goyal, Du, Joshi, Chen, Levy, Lewis,
  Zettlemoyer, and Stoyanov]{liu2019roberta}
Yinhan Liu, Myle Ott, Naman Goyal, Jingfei Du, Mandar Joshi, Danqi Chen, Omer
  Levy, Mike Lewis, Luke Zettlemoyer, and Veselin Stoyanov.
\newblock Roberta: A robustly optimized bert pretraining approach.
\newblock \emph{arXiv preprint arXiv:1907.11692}, 2019.

\bibitem[Machado et~al.(2018)Machado, Bellemare, Talvitie, Veness, Hausknecht,
  and Bowling]{machado2018revisiting}
Marlos~C Machado, Marc~G Bellemare, Erik Talvitie, Joel Veness, Matthew
  Hausknecht, and Michael Bowling.
\newblock Revisiting the arcade learning environment: Evaluation protocols and
  open problems for general agents.
\newblock \emph{Journal of Artificial Intelligence Research}, 61:\penalty0
  523--562, 2018.

\bibitem[Marie et~al.(2021)Marie, Fujita, and Rubino]{benjaminccientific}
Benjamin Marie, Atsushi Fujita, and Raphael Rubino.
\newblock Scientific credibility of machine translation research: A
  meta-evaluation of 769 papers.
\newblock \emph{arXiv preprint arXiv:2106.15195}, 2021.

\bibitem[Martin et~al.(2017)Martin, Sasikumar, Everitt, and
  Hutter]{martin2017count}
Jarryd Martin, Suraj~Narayanan Sasikumar, Tom Everitt, and Marcus Hutter.
\newblock Count-based exploration in feature space for reinforcement learning.
\newblock \emph{arXiv preprint arXiv:1706.08090}, 2017.

\bibitem[Metzler and Kurland(2012)]{metzler2012experimental}
Donald Metzler and Oren Kurland.
\newblock Experimental methods for information retrieval.
\newblock In \emph{Proceedings of the 35th international ACM SIGIR conference
  on Research and development in information retrieval}, pages 1185--1186,
  2012.

\bibitem[Mikolov et~al.(2013)Mikolov, Sutskever, Chen, Corrado, and
  Dean]{mikolov2013distributed}
Tomas Mikolov, Ilya Sutskever, Kai Chen, Greg~S Corrado, and Jeff Dean.
\newblock Distributed representations of words and phrases and their
  compositionality.
\newblock In \emph{Advances in neural information processing systems}, pages
  3111--3119, 2013.

\bibitem[Mishra and Arunkumar(2021)]{mishra2021robust}
Swaroop Mishra and Anjana Arunkumar.
\newblock How robust are model rankings: A leaderboard customization approach
  for equitable evaluation.
\newblock \emph{arXiv preprint arXiv:2106.05532}, 2021.

\bibitem[Mnih et~al.(2013)Mnih, Kavukcuoglu, Silver, Graves, Antonoglou,
  Wierstra, and Riedmiller]{mnih2013playing}
Volodymyr Mnih, Koray Kavukcuoglu, David Silver, Alex Graves, Ioannis
  Antonoglou, Daan Wierstra, and Martin Riedmiller.
\newblock Playing atari with deep reinforcement learning.
\newblock \emph{arXiv preprint arXiv:1312.5602}, 2013.

\bibitem[Mnih et~al.(2015)Mnih, Kavukcuoglu, Silver, Rusu, Veness, Bellemare,
  Graves, Riedmiller, Fidjeland, Ostrovski, et~al.]{mnih2015human}
Volodymyr Mnih, Koray Kavukcuoglu, David Silver, Andrei~A Rusu, Joel Veness,
  Marc~G Bellemare, Alex Graves, Martin Riedmiller, Andreas~K Fidjeland, Georg
  Ostrovski, et~al.
\newblock Human-level control through deep reinforcement learning.
\newblock \emph{nature}, 518\penalty0 (7540):\penalty0 529--533, 2015.

\bibitem[Mnih et~al.(2016)Mnih, Badia, Mirza, Graves, Lillicrap, Harley,
  Silver, and Kavukcuoglu]{mnih2016asynchronous}
Volodymyr Mnih, Adria~Puigdomenech Badia, Mehdi Mirza, Alex Graves, Timothy
  Lillicrap, Tim Harley, David Silver, and Koray Kavukcuoglu.
\newblock Asynchronous methods for deep reinforcement learning.
\newblock In \emph{International conference on machine learning}, pages
  1928--1937. PMLR, 2016.

\bibitem[Musgrave et~al.(2020)Musgrave, Belongie, and Lim]{musgrave2020metric}
Kevin Musgrave, Serge Belongie, and Ser-Nam Lim.
\newblock A metric learning reality check.
\newblock In \emph{European Conference on Computer Vision}, pages 681--699.
  Springer, 2020.

\bibitem[Nair et~al.(2015)Nair, Srinivasan, Blackwell, Alcicek, Fearon,
  De~Maria, Panneershelvam, Suleyman, Beattie, Petersen,
  et~al.]{nair2015massively}
Arun Nair, Praveen Srinivasan, Sam Blackwell, Cagdas Alcicek, Rory Fearon,
  Alessandro De~Maria, Vedavyas Panneershelvam, Mustafa Suleyman, Charles
  Beattie, Stig Petersen, et~al.
\newblock Massively parallel methods for deep reinforcement learning.
\newblock \emph{arXiv preprint arXiv:1507.04296}, 2015.

\bibitem[Narang et~al.(2021)Narang, Chung, Tay, Fedus, Fevry, Matena, Malkan,
  Fiedel, Shazeer, Lan, Zhou, Li, Ding, Marcus, Roberts, and
  Raffel]{narang2021transformer}
Sharan Narang, Hyung~Won Chung, Yi~Tay, William Fedus, Thibault Fevry, Michael
  Matena, Karishma Malkan, Noah Fiedel, Noam Shazeer, Zhenzhong Lan, Yanqi
  Zhou, Wei Li, Nan Ding, Jake Marcus, Adam Roberts, and Colin Raffel.
\newblock Do transformer modifications transfer across implementations and
  applications?, 2021.

\bibitem[Netzer et~al.(2011)Netzer, Wang, Coates, Bissacco, Wu, and
  Ng]{netzer2011reading}
Yuval Netzer, Tao Wang, Adam Coates, Alessandro Bissacco, Bo~Wu, and Andrew~Y
  Ng.
\newblock Reading digits in natural images with unsupervised feature learning.
\newblock 2011.

\bibitem[Nilsback and Zisserman(2008)]{nilsback2008automated}
Maria-Elena Nilsback and Andrew Zisserman.
\newblock Automated flower classification over a large number of classes.
\newblock In \emph{2008 Sixth Indian Conference on Computer Vision, Graphics \&
  Image Processing}, pages 722--729. IEEE, 2008.

\bibitem[Northcutt et~al.(2021)Northcutt, Athalye, and
  Mueller]{northcutt2021pervasive}
Curtis~G Northcutt, Anish Athalye, and Jonas Mueller.
\newblock Pervasive label errors in test sets destabilize machine learning
  benchmarks.
\newblock \emph{arXiv preprint arXiv:2103.14749}, 2021.

\bibitem[Parkhi et~al.(2012)Parkhi, Vedaldi, Zisserman, and
  Jawahar]{parkhi2012cats}
Omkar~M Parkhi, Andrea Vedaldi, Andrew Zisserman, and CV~Jawahar.
\newblock Cats and dogs.
\newblock In \emph{2012 IEEE conference on computer vision and pattern
  recognition}, pages 3498--3505. IEEE, 2012.

\bibitem[Pei et~al.(2019)Pei, Zhang, Zhang, Sun, Lin, Sun, Wu, Jiang, Ge, Ou,
  et~al.]{pei2019personalized}
Changhua Pei, Yi~Zhang, Yongfeng Zhang, Fei Sun, Xiao Lin, Hanxiao Sun, Jian
  Wu, Peng Jiang, Junfeng Ge, Wenwu Ou, et~al.
\newblock Personalized re-ranking for recommendation.
\newblock In \emph{Proceedings of the 13th ACM Conference on Recommender
  Systems}, pages 3--11, 2019.

\bibitem[Perazzi et~al.(2016)Perazzi, Pont-Tuset, McWilliams, Van~Gool, Gross,
  and Sorkine-Hornung]{perazzi2016benchmark}
Federico Perazzi, Jordi Pont-Tuset, Brian McWilliams, Luc Van~Gool, Markus
  Gross, and Alexander Sorkine-Hornung.
\newblock A benchmark dataset and evaluation methodology for video object
  segmentation.
\newblock In \emph{Proceedings of the IEEE conference on computer vision and
  pattern recognition}, pages 724--732, 2016.

\bibitem[Ponce et~al.(2006)Ponce, Berg, Everingham, Forsyth, Hebert, Lazebnik,
  Marszalek, Schmid, Russell, Torralba, et~al.]{ponce2006dataset}
Jean Ponce, Tamara~L Berg, Mark Everingham, David~A Forsyth, Martial Hebert,
  Svetlana Lazebnik, Marcin Marszalek, Cordelia Schmid, Bryan~C Russell,
  Antonio Torralba, et~al.
\newblock Dataset issues in object recognition.
\newblock In \emph{Toward category-level object recognition}, pages 29--48.
  Springer, 2006.

\bibitem[Pritzel et~al.(2017)Pritzel, Uria, Srinivasan, Badia, Vinyals,
  Hassabis, Wierstra, and Blundell]{pritzel2017neural}
Alexander Pritzel, Benigno Uria, Sriram Srinivasan, Adria~Puigdomenech Badia,
  Oriol Vinyals, Demis Hassabis, Daan Wierstra, and Charles Blundell.
\newblock Neural episodic control.
\newblock In \emph{International Conference on Machine Learning}, pages
  2827--2836. PMLR, 2017.

\bibitem[Rajpurkar et~al.(2016)Rajpurkar, Zhang, Lopyrev, and
  Liang]{rajpurkar2016squad}
Pranav Rajpurkar, Jian Zhang, Konstantin Lopyrev, and Percy Liang.
\newblock Squad: 100,000+ questions for machine comprehension of text.
\newblock \emph{arXiv preprint arXiv:1606.05250}, 2016.

\bibitem[Rajpurkar et~al.(2018)Rajpurkar, Jia, and Liang]{rajpurkar2018know}
Pranav Rajpurkar, Robin Jia, and Percy Liang.
\newblock Know what you don't know: Unanswerable questions for squad.
\newblock \emph{arXiv preprint arXiv:1806.03822}, 2018.

\bibitem[Recht et~al.(2018)Recht, Roelofs, Schmidt, and
  Shankar]{recht2018cifar}
Benjamin Recht, Rebecca Roelofs, Ludwig Schmidt, and Vaishaal Shankar.
\newblock Do cifar-10 classifiers generalize to cifar-10?
\newblock \emph{arXiv preprint arXiv:1806.00451}, 2018.

\bibitem[Recht et~al.(2019)Recht, Roelofs, Schmidt, and
  Shankar]{recht2019imagenet}
Benjamin Recht, Rebecca Roelofs, Ludwig Schmidt, and Vaishaal Shankar.
\newblock Do {ImageNet} classifiers generalize to {ImageNet}?
\newblock In \emph{International Conference on Machine Learning}, pages
  5389--5400. PMLR, 2019.

\bibitem[Rendle et~al.(2020)Rendle, Krichene, Zhang, and
  Anderson]{rendle2020neural}
Steffen Rendle, Walid Krichene, Li~Zhang, and John Anderson.
\newblock Neural collaborative filtering vs. matrix factorization revisited.
\newblock In \emph{Fourteenth ACM Conference on Recommender Systems}, pages
  240--248, 2020.

\bibitem[Roelofs et~al.(2019)Roelofs, Shankar, Recht, Fridovich-Keil, Hardt,
  Miller, and Schmidt]{roelofs2019meta}
Rebecca Roelofs, Vaishaal Shankar, Benjamin Recht, Sara Fridovich-Keil, Moritz
  Hardt, John Miller, and Ludwig Schmidt.
\newblock A meta-analysis of overfitting in machine learning.
\newblock \emph{Advances in Neural Information Processing Systems},
  32:\penalty0 9179--9189, 2019.

\bibitem[Schoch et~al.(2020)Schoch, Yang, and Ji]{schoch2020problem}
Stephanie Schoch, Diyi Yang, and Yangfeng Ji.
\newblock “this is a problem, don’t you agree?” framing and bias in human
  evaluation for natural language generation.
\newblock In \emph{Proceedings of the 1st Workshop on Evaluating NLG
  Evaluation}, pages 10--16, 2020.

\bibitem[Schrittwieser et~al.(2021)Schrittwieser, Hubert, Mandhane, Barekatain,
  Antonoglou, and Silver]{schrittwieser2021online}
Julian Schrittwieser, Thomas Hubert, Amol Mandhane, Mohammadamin Barekatain,
  Ioannis Antonoglou, and David Silver.
\newblock Online and offline reinforcement learning by planning with a learned
  model.
\newblock \emph{arXiv preprint arXiv:2104.06294}, 2021.

\bibitem[Sculley et~al.(2018)Sculley, Snoek, Wiltschko, and
  Rahimi]{sculley2018winner}
David Sculley, Jasper Snoek, Alex Wiltschko, and Ali Rahimi.
\newblock Winner's curse? on pace, progress, and empirical rigor.
\newblock 2018.

\bibitem[Seo et~al.(2018)Seo, Kwiatkowski, Parikh, Farhadi, and
  Hajishirzi]{seo-etal-2018-phrase}
Minjoon Seo, Tom Kwiatkowski, Ankur Parikh, Ali Farhadi, and Hannaneh
  Hajishirzi.
\newblock Phrase-indexed question answering: A new challenge for scalable
  document comprehension.
\newblock In \emph{Proceedings of the 2018 Conference on Empirical Methods in
  Natural Language Processing}, pages 559--564, Brussels, Belgium,
  October-November 2018. Association for Computational Linguistics.
\newblock \doi{10.18653/v1/D18-1052}.
\newblock URL \url{https://www.aclweb.org/anthology/D18-1052}.

\bibitem[Silver et~al.(2016)Silver, Huang, Maddison, Guez, Sifre, Van
  Den~Driessche, Schrittwieser, Antonoglou, Panneershelvam, Lanctot,
  et~al.]{silver2016mastering}
David Silver, Aja Huang, Chris~J Maddison, Arthur Guez, Laurent Sifre, George
  Van Den~Driessche, Julian Schrittwieser, Ioannis Antonoglou, Veda
  Panneershelvam, Marc Lanctot, et~al.
\newblock Mastering the game of go with deep neural networks and tree search.
\newblock \emph{nature}, 529\penalty0 (7587):\penalty0 484--489, 2016.

\bibitem[So et~al.(2019)So, Le, and Liang]{so2019evolved}
David So, Quoc Le, and Chen Liang.
\newblock The evolved transformer.
\newblock In \emph{International Conference on Machine Learning}, pages
  5877--5886. PMLR, 2019.

\bibitem[Socher et~al.(2013)Socher, Perelygin, Wu, Chuang, Manning, Ng, and
  Potts]{socher-etal-2013-recursive}
Richard Socher, Alex Perelygin, Jean Wu, Jason Chuang, Christopher~D. Manning,
  Andrew Ng, and Christopher Potts.
\newblock Recursive deep models for semantic compositionality over a sentiment
  treebank.
\newblock In \emph{Proceedings of the 2013 Conference on Empirical Methods in
  Natural Language Processing}, pages 1631--1642, Seattle, Washington, USA,
  October 2013. Association for Computational Linguistics.
\newblock URL \url{https://www.aclweb.org/anthology/D13-1170}.

\bibitem[Steiner et~al.(2021)Steiner, Kolesnikov, Zhai, Wightman, Uszkoreit,
  and Beyer]{steiner2021train}
Andreas Steiner, Alexander Kolesnikov, Xiaohua Zhai, Ross Wightman, Jakob
  Uszkoreit, and Lucas Beyer.
\newblock How to train your vit? data, augmentation, and regularization in
  vision transformers.
\newblock \emph{arXiv preprint arXiv:2106.10270}, 2021.

\bibitem[Sutskever et~al.(2014)Sutskever, Vinyals, and
  Le]{sutskever2014sequence}
Ilya Sutskever, Oriol Vinyals, and Quoc~V Le.
\newblock Sequence to sequence learning with neural networks.
\newblock \emph{arXiv preprint arXiv:1409.3215}, 2014.

\bibitem[Swaminathan et~al.(2016)Swaminathan, Krishnamurthy, Agarwal,
  Dud{\'\i}k, Langford, Jose, and Zitouni]{swaminathan2016off}
Adith Swaminathan, Akshay Krishnamurthy, Alekh Agarwal, Miroslav Dud{\'\i}k,
  John Langford, Damien Jose, and Imed Zitouni.
\newblock Off-policy evaluation for slate recommendation.
\newblock \emph{arXiv preprint arXiv:1605.04812}, 2016.

\bibitem[Tabrizi et~al.(2015)Tabrizi, Dadashkarimi, Dehghani, Nasr~Esfahani,
  and Shakery]{tabrizi2015revisiting}
Shayan~A Tabrizi, Javid Dadashkarimi, Mostafa Dehghani, Hassan Nasr~Esfahani,
  and Azadeh Shakery.
\newblock Revisiting optimal rank aggregation: A dynamic programming approach.
\newblock In \emph{Proceedings of the 2015 International Conference on The
  Theory of Information Retrieval}, pages 353--356, 2015.

\bibitem[Tassa et~al.(2018)Tassa, Doron, Muldal, Erez, Li, Casas, Budden,
  Abdolmaleki, Merel, Lefrancq, et~al.]{tassa2018deepmind}
Yuval Tassa, Yotam Doron, Alistair Muldal, Tom Erez, Yazhe Li, Diego de~Las
  Casas, David Budden, Abbas Abdolmaleki, Josh Merel, Andrew Lefrancq, et~al.
\newblock Deepmind control suite.
\newblock \emph{arXiv preprint arXiv:1801.00690}, 2018.

\bibitem[Tay et~al.(2020{\natexlab{a}})Tay, Bahri, Metzler, Juan, Zhao, and
  Zheng]{tay2020synthesizer}
Yi~Tay, Dara Bahri, Donald Metzler, Da-Cheng Juan, Zhe Zhao, and Che Zheng.
\newblock Synthesizer: Rethinking self-attention in transformer models.
\newblock \emph{arXiv preprint arXiv:2005.00743}, 2020{\natexlab{a}}.

\bibitem[Tay et~al.(2020{\natexlab{b}})Tay, Dehghani, Abnar, Shen, Bahri, Pham,
  Rao, Yang, Ruder, and Metzler]{tay2020long}
Yi~Tay, Mostafa Dehghani, Samira Abnar, Yikang Shen, Dara Bahri, Philip Pham,
  Jinfeng Rao, Liu Yang, Sebastian Ruder, and Donald Metzler.
\newblock Long range arena: A benchmark for efficient transformers.
\newblock \emph{arXiv preprint arXiv:2011.04006}, 2020{\natexlab{b}}.

\bibitem[Tay et~al.(2020{\natexlab{c}})Tay, Dehghani, Bahri, and
  Metzler]{tay2020efficient}
Yi~Tay, Mostafa Dehghani, Dara Bahri, and Donald Metzler.
\newblock Efficient transformers: A survey.
\newblock \emph{arXiv preprint arXiv:2009.06732}, 2020{\natexlab{c}}.

\bibitem[Tay et~al.(2021)Tay, Dehghani, Gupta, Bahri, Aribandi, Qin, and
  Metzler]{tay2021pre}
Yi~Tay, Mostafa Dehghani, Jai Gupta, Dara Bahri, Vamsi Aribandi, Zhen Qin, and
  Donald Metzler.
\newblock Are pre-trained convolutions better than pre-trained transformers?
\newblock \emph{arXiv preprint arXiv:2105.03322}, 2021.

\bibitem[Tolstikhin et~al.(2017)Tolstikhin, Bousquet, Gelly, and
  Schoelkopf]{tolstikhin2017wasserstein}
Ilya Tolstikhin, Olivier Bousquet, Sylvain Gelly, and Bernhard Schoelkopf.
\newblock Wasserstein auto-encoders.
\newblock \emph{arXiv preprint arXiv:1711.01558}, 2017.

\bibitem[Torralba and Efros(2011)]{torralba2011unbiased}
Antonio Torralba and Alexei~A Efros.
\newblock Unbiased look at dataset bias.
\newblock In \emph{CVPR 2011}, pages 1521--1528. IEEE, 2011.

\bibitem[Tran et~al.(2018)Tran, Bisazza, and Monz]{tran-etal-2018-importance}
Ke~Tran, Arianna Bisazza, and Christof Monz.
\newblock The importance of being recurrent for modeling hierarchical
  structure.
\newblock In \emph{Proceedings of the 2018 Conference on Empirical Methods in
  Natural Language Processing}, 2018.

\bibitem[Tschannen et~al.(2020)Tschannen, Djolonga, Ritter, Mahendran, Houlsby,
  Gelly, and Lucic]{tschannen2020self}
Michael Tschannen, Josip Djolonga, Marvin Ritter, Aravindh Mahendran, Neil
  Houlsby, Sylvain Gelly, and Mario Lucic.
\newblock Self-supervised learning of video-induced visual invariances.
\newblock In \emph{Proceedings of the IEEE/CVF Conference on Computer Vision
  and Pattern Recognition}, pages 13806--13815, 2020.

\bibitem[Turing(1950)]{Turing1950}
Alan~M. Turing.
\newblock {Computing machinery and intelligence}.
\newblock \emph{Mind}, LIX\penalty0 (236):\penalty0 433--460, 1950.

\bibitem[Van Der~Lee et~al.(2019)Van Der~Lee, Gatt, Van~Miltenburg, Wubben, and
  Krahmer]{van2019best}
Chris Van Der~Lee, Albert Gatt, Emiel Van~Miltenburg, Sander Wubben, and Emiel
  Krahmer.
\newblock Best practices for the human evaluation of automatically generated
  text.
\newblock In \emph{Proceedings of the 12th International Conference on Natural
  Language Generation}, pages 355--368, 2019.

\bibitem[Van~Hasselt et~al.(2016)Van~Hasselt, Guez, and Silver]{van2016deep}
Hado Van~Hasselt, Arthur Guez, and David Silver.
\newblock Deep reinforcement learning with double q-learning.
\newblock In \emph{Proceedings of the AAAI Conference on Artificial
  Intelligence}, volume~30, 2016.

\bibitem[Vania et~al.(2020)Vania, Htut, Huang, Mungra, Yuanzhe~Pang, Phang,
  Liu, Cho, and Bowman]{vania2020comparing}
Clara Vania, Phu~Mon Htut, William Huang, Dhara Mungra, Richard Yuanzhe~Pang,
  Jason Phang, Haokun Liu, Kyunghyun Cho, and Samuel~R. Bowman.
\newblock Comparing test sets with item response theory.
\newblock \emph{arXiv preprint arXiv:2106.00840}, 2020.

\bibitem[Vaswani et~al.(2017)Vaswani, Shazeer, Parmar, Uszkoreit, Jones, Gomez,
  Kaiser, and Polosukhin]{vaswani2017attention}
Ashish Vaswani, Noam Shazeer, Niki Parmar, Jakob Uszkoreit, Llion Jones,
  Aidan~N Gomez, {\L}ukasz Kaiser, and Illia Polosukhin.
\newblock Attention is all you need.
\newblock In \emph{Advances in neural information processing systems}, pages
  5998--6008, 2017.

\bibitem[Veeling et~al.(2018)Veeling, Linmans, Winkens, Cohen, and
  Welling]{veeling2018rotation}
Bastiaan~S Veeling, Jasper Linmans, Jim Winkens, Taco Cohen, and Max Welling.
\newblock Rotation equivariant cnns for digital pathology.
\newblock In \emph{International Conference on Medical image computing and
  computer-assisted intervention}, pages 210--218. Springer, 2018.

\bibitem[Wang et~al.(2018)Wang, Singh, Michael, Hill, Levy, and
  Bowman]{wang-etal-2018-glue}
Alex Wang, Amanpreet Singh, Julian Michael, Felix Hill, Omer Levy, and Samuel
  Bowman.
\newblock {GLUE}: A multi-task benchmark and analysis platform for natural
  language understanding.
\newblock In \emph{Proceedings of the 2018 {EMNLP} Workshop {B}lackbox{NLP}:
  Analyzing and Interpreting Neural Networks for {NLP}}, pages 353--355,
  Brussels, Belgium, November 2018. Association for Computational Linguistics.
\newblock \doi{10.18653/v1/W18-5446}.
\newblock URL \url{https://www.aclweb.org/anthology/W18-5446}.

\bibitem[Wang et~al.(2019)Wang, Pruksachatkun, Nangia, Singh, Michael, Hill,
  Levy, and Bowman]{wang2019superglue}
Alex Wang, Yada Pruksachatkun, Nikita Nangia, Amanpreet Singh, Julian Michael,
  Felix Hill, Omer Levy, and Samuel~R Bowman.
\newblock Superglue: A stickier benchmark for general-purpose language
  understanding systems.
\newblock \emph{arXiv preprint arXiv:1905.00537}, 2019.

\bibitem[Wang et~al.(2016)Wang, Schaul, Hessel, Hasselt, Lanctot, and
  Freitas]{wang2016dueling}
Ziyu Wang, Tom Schaul, Matteo Hessel, Hado Hasselt, Marc Lanctot, and Nando
  Freitas.
\newblock Dueling network architectures for deep reinforcement learning.
\newblock In \emph{International conference on machine learning}, pages
  1995--2003. PMLR, 2016.

\bibitem[Williams et~al.(2017)Williams, Nangia, and Bowman]{williams2017broad}
Adina Williams, Nikita Nangia, and Samuel~R Bowman.
\newblock A broad-coverage challenge corpus for sentence understanding through
  inference.
\newblock \emph{arXiv preprint arXiv:1704.05426}, 2017.

\bibitem[Xiao et~al.(2010)Xiao, Hays, Ehinger, Oliva, and
  Torralba]{xiao2010sun}
Jianxiong Xiao, James Hays, Krista~A Ehinger, Aude Oliva, and Antonio Torralba.
\newblock Sun database: Large-scale scene recognition from abbey to zoo.
\newblock In \emph{2010 IEEE computer society conference on computer vision and
  pattern recognition}, pages 3485--3492. IEEE, 2010.

\bibitem[Yadav and Bottou(2019)]{yadav2019cold}
Chhavi Yadav and L{\'e}on Bottou.
\newblock Cold case: The lost mnist digits.
\newblock \emph{arXiv preprint arXiv:1905.10498}, 2019.

\bibitem[Yang et~al.(2019)Yang, Dai, Yang, Carbonell, Salakhutdinov, and
  Le]{yang2019xlnet}
Zhilin Yang, Zihang Dai, Yiming Yang, Jaime Carbonell, Russ~R Salakhutdinov,
  and Quoc~V Le.
\newblock Xlnet: Generalized autoregressive pretraining for language
  understanding.
\newblock \emph{Advances in neural information processing systems}, 32, 2019.

\bibitem[Yang et~al.(2020)Yang, Dai, Yang, Carbonell, Salakhutdinov, and
  Le]{yang2020xlnet}
Zhilin Yang, Zihang Dai, Yiming Yang, Jaime Carbonell, Ruslan Salakhutdinov,
  and Quoc~V. Le.
\newblock Xlnet: Generalized autoregressive pretraining for language
  understanding, 2020.

\bibitem[Yi et~al.(2019)Yi, Yang, Hong, Cheng, Heldt, Kumthekar, Zhao, Wei, and
  Chi]{yi2019sampling}
Xinyang Yi, Ji~Yang, Lichan Hong, Derek~Zhiyuan Cheng, Lukasz Heldt, Aditee
  Kumthekar, Zhe Zhao, Li~Wei, and Ed~Chi.
\newblock Sampling-bias-corrected neural modeling for large corpus item
  recommendations.
\newblock In \emph{Proceedings of the 13th ACM Conference on Recommender
  Systems}, pages 269--277, 2019.

\bibitem[Zhai et~al.(2019)Zhai, Puigcerver, Kolesnikov, Ruyssen, Riquelme,
  Lucic, Djolonga, Pinto, Neumann, Dosovitskiy, et~al.]{zhai2019large}
Xiaohua Zhai, Joan Puigcerver, Alexander Kolesnikov, Pierre Ruyssen, Carlos
  Riquelme, Mario Lucic, Josip Djolonga, Andre~Susano Pinto, Maxim Neumann,
  Alexey Dosovitskiy, et~al.
\newblock A large-scale study of representation learning with the visual task
  adaptation benchmark.
\newblock \emph{arXiv preprint arXiv:1910.04867}, 2019.

\bibitem[Zhang et~al.(2019)Zhang, Yao, Sun, and Tay]{zhang2019deep}
Shuai Zhang, Lina Yao, Aixin Sun, and Yi~Tay.
\newblock Deep learning based recommender system: A survey and new
  perspectives.
\newblock \emph{ACM Computing Surveys (CSUR)}, 52\penalty0 (1):\penalty0 1--38,
  2019.

\bibitem[Zhang(2021)]{covariant-robotics-benchmark}
Tianhao Zhang.
\newblock The need for performance based assessments, 2021.
\newblock URL \url{https://covariant.ai/news/performance-based-assessments}.

\bibitem[Zhang et~al.(2020)Zhang, Cheng, Yao, Yi, Hong, and
  Chi]{zhang2020model}
Yin Zhang, Derek~Zhiyuan Cheng, Tiansheng Yao, Xinyang Yi, Lichan Hong, and
  Ed~H Chi.
\newblock A model of two tales: Dual transfer learning framework for improved
  long-tail item recommendation.
\newblock \emph{arXiv preprint arXiv:2010.15982}, 2020.

\bibitem[Zhao et~al.(2019)Zhao, Hong, Wei, Chen, Nath, Andrews, Kumthekar,
  Sathiamoorthy, Yi, and Chi]{zhao2019recommending}
Zhe Zhao, Lichan Hong, Li~Wei, Jilin Chen, Aniruddh Nath, Shawn Andrews, Aditee
  Kumthekar, Maheswaran Sathiamoorthy, Xinyang Yi, and Ed~Chi.
\newblock Recommending what video to watch next: a multitask ranking system.
\newblock In \emph{Proceedings of the 13th ACM Conference on Recommender
  Systems}, pages 43--51, 2019.

\bibitem[Zheng et~al.(2019)Zheng, Lu, He, Xie, He, Li, Noroozi, Dong, and
  Philip]{zheng2019mars}
Lei Zheng, Chun-Ta Lu, Lifang He, Sihong Xie, Huang He, Chaozhuo Li, Vahid
  Noroozi, Bowen Dong, and S~Yu Philip.
\newblock Mars: Memory attention-aware recommender system.
\newblock In \emph{2019 IEEE International Conference on Data Science and
  Advanced Analytics (DSAA)}, pages 11--20. IEEE, 2019.

\end{thebibliography}
